\documentclass[runningheads]{llncs}
\usepackage[T1]{fontenc}
\usepackage{multirow}
\usepackage[title]{appendix}
\usepackage{graphicx}
\usepackage{wrapfig}
\usepackage{blindtext}
\usepackage{amsmath}
\usepackage{amssymb}
\usepackage{booktabs}
\usepackage{siunitx}
\usepackage{subcaption}
\usepackage[misc,geometry]{ifsym}
\usepackage[hidelinks,colorlinks=true,allcolors=blue]{hyperref}
\usepackage{color}
\usepackage{algorithm,algpseudocode}

\newcommand{\etal}{\textit{et al.}}
\newcommand{\rvlcdip}{RVL-CDIP~\cite{harley2015-rvlcdip}}
\newcommand{\doclaynet}{DocLayNet~\cite{doclaynet}}
\newcommand{\tobacco}{Tobacco3482}
\hypersetup{final}

\newcommand{\ldce}{LDCE~\cite{farid2023latentdiffusioncounterfactualexplanations}}
\newcommand{\sldce}{S-LDCE (ours)}
\newcommand{\ldcehpr}{LDCE+HPR (ours)}
\newcommand{\sldcehpr}{DocVCE (S-LDCE+HPR) (ours)}

\makeatletter

\renewcommand\section{\@startsection{section}{1}{\z@}%
	{-8\p@ \@plus -4\p@ \@minus -4\p@}%
	{6\p@ \@plus 4\p@ \@minus 4\p@}%
	{\normalfont\large\bfseries\boldmath
		\rightskip=\z@ \@plus 8em\pretolerance=10000 }}
\renewcommand\subsection{\@startsection{subsection}{2}{\z@}%
	{-8\p@ \@plus -4\p@ \@minus -4\p@}%
	{6\p@ \@plus 4\p@ \@minus 4\p@}%
	{\normalfont\normalsize\bfseries\boldmath
		\rightskip=\z@ \@plus 8em\pretolerance=10000 }}
\renewcommand\subsubsection{\@startsection{subsubsection}{3}{\z@}%
	{-8\p@ \@plus -4\p@ \@minus -4\p@}%
	{-0.5em \@plus -0.22em \@minus -0.1em}%
	{\normalfont\normalsize\bfseries\boldmath}}
\makeatother

\begin{document}
\title{
	DocVCE: Diffusion-based Visual Counterfactual Explanations for Document Image Classification
}
\titlerunning{DocVCE: Counterfactual Explanations for Document Image Classification}
\author{
	Saifullah Saifullah(\Letter)\inst{1,2} \orcidID{0000-0003-3098-2458} \and
	Stefan Agne\inst{1,3} \orcidID{0000-0002-9697-4285} \and
	Andreas Dengel\inst{1,2} \orcidID{0000-0002-6100-8255} \and
	Sheraz Ahmed\inst{1,3} \orcidID{0000-0002-4239-6520}}
\authorrunning{S. Saifullah \textit{et al.}}
\institute{
	Smarte Daten and Wissensdienste (SDS), Deutsches Forschungszentrum für Künstliche Intelligenz GmbH (DFKI), Trippstadter Straße 122,
	67663 Kaiserslautern, Germany\\\email{\{firstname.lastname\}@dfki.de}\\ \and
	Department of Computer Science, RPTU Kaiserslautern-Landau, Erwin-Schrödinger-Straße 52, 67663 Kaiserslautern, Germany\and
	DeepReader GmbH, 67663 Kaiserlautern, Germany\\
}
\maketitle
\begin{abstract}
	As black-box AI-driven decision-making systems become increasingly widespread in modern document processing workflows, improving their transparency and reliability has become critical, especially in high-stakes applications where biases or spurious correlations in decision-making could lead to serious consequences.
    One vital component often found in such document processing workflows is document image classification, which, despite its widespread use, remains difficult to explain.
	While some recent works have attempted to explain the decisions of document image classification models through feature-importance maps, these maps are often difficult to interpret and fail to provide insights into the global features learned by the model.
	In this paper, we aim to bridge this research gap by introducing generative document counterfactuals that provide meaningful insights into the model's decision-making through actionable explanations.
	In particular, we propose DocVCE, a novel approach that leverages latent diffusion models in combination with classifier guidance to first generate plausible in-distribution visual counterfactual explanations, and then performs hierarchical patch-wise refinement to search for a refined counterfactual that is closest to the target factual image.
	We demonstrate the effectiveness of our approach through a rigorous qualitative and quantitative assessment on 3 different document classification datasets---RVL-CDIP, Tobacco3482, and DocLayNet---and 3 different models---ResNet, ConvNeXt, and DiT---using well-established evaluation criteria such as validity, closeness, and realism.
	To the best of the authors' knowledge, this is the first work to explore generative counterfactual explanations in document image analysis.
    The source code for this work will be made publicly available upon acceptance.
	\keywords{Counterfactual Explanations \and Explainable Document Classification \and Explainable AI \and Document Image Classification \and  Model Interpretability \and}
\end{abstract}
\section{Introduction}
\label{sec:introduction}
As black-box deep learning (DL)-based information extraction models continue to achieve remarkable performance breakthroughs across a wide variety of document analysis tasks~\cite{Saifullah2022-docxclassifier,layoutlmv3,shen2021layoutparser,tilt}, a primary concern that still hinders their full integration into modern automated workflows---especially in high-stakes scenarios such as legal or law enforcement applications---is their inherent lack of transparency~\cite{safety-ml,rudin2019stop,wexler_jail_sentences} and reliability~\cite{intro-adversarial-2,Hendrycks2019BenchmarkingNN,saifullah-robustness}.
This becomes especially relevant in light of recent regulations, such as the GDPR~\cite{GDPR2016a} and the AI Act 2022, which impose legal requirements on these systems to be both transparent and accountable for their decisions.

Several recent studies have shown that most black-box deep learning models are not only susceptible to making unfair and biased decisions~\cite{biases-survey,geirhos2022imagenettrained,lucieri-shape-bias,hooker2020characterising,rudin2019stop}---typically stemming from hidden biases in the training data---but also exhibit vulnerabilities when faced with out-of-distribution data at test time~\cite{intro-adversarial-2,Hendrycks2019BenchmarkingNN,saifullah-robustness}. For instance, a recent work by Saifullah~\etal~\cite{Saifullah2023-docxai} highlighted the spurious correlations learned by the black-box DL models on the document classification task. Similarly, recently introduced document robustness benchmarks, such as RVL-CDIP-D~\cite{saifullah-robustness} or RoDLA~\cite{chen2024rodla} have demonstrated the vulnerability of state-of-the-art document image analysis models to minor input data perturbations. Such failure modes have given rise to the field of eXplainable AI (XAI) which aims to make these models more transparent and accountable.

A wide variety of explainability methods~\cite{intro-survey-1,intro-survey-2} have been proposed in recent years to interpret and validate the decisions and behaviors of deep learning models~\cite{lime,grad-cam,smooth-grad,shap,saliency,nemirovsky2020countergan}, the majority of which can be broadly classified into two main categories: the attribution-based methods~\cite{lime,grad-cam,smooth-grad,shap,saliency} and the generative counterfactual approaches~\cite{nemirovsky2020countergan,jeanneret2022diffusion}.
While several recent works have extended attribution-based methods to various document analysis tasks~\cite{Saifullah2023-docxai,Brini2022,pouliquen2024weaklysupervisedtraininghologram} with promising results, the explanations produced by these methods are often difficult to interpret and fail to provide additional information on whether the model's decisions are influenced by global typographical and structural characteristics, such as font styles, text formatting, or document layout.
On the other hand, generative counterfactual explanations~\cite{farid2023latentdiffusioncounterfactualexplanations,Augustin2022Diffusion,jeanneret2022diffusion} have recently shown great promise in uncovering the influence of such abstract features in natural image classification.
In particular, diffusion models~\cite{dhariwal2021diffusionmodelsbeatgans} have recently gained considerable attention for generating counterfactual explanations~\cite{farid2023latentdiffusioncounterfactualexplanations,Augustin2022Diffusion,jeanneret2022diffusion} due to their high-quality generative capabilities. However, while diffusion has also been widely explored in recent years for synthetic document image generation~\cite{ldms-doc,hamdani-doc,guan2024idnetnoveldatasetidentity}, we found no existing research that specifically focuses on diffusion-based counterfactual explanations for document image analysis.

In this work, we explore the potential of generative visual counterfactual explanations for interpreting the decisions of deep-learning based document image classification models.
In particular, we introduce diffusion-based visual counterfactual explanations for document image classification (DocVCE), a novel approach that first uses latent diffusion models~\cite{rombach2022highresolutionimagesynthesislatent} in combination with classifier-guided stochastic sampling~\cite{ho2022classifierfreediffusionguidance,dhariwal2021diffusionmodelsbeatgans} to generate high-level, in-distribution counterfactual images, and then performs hierarchical patch-wise refinement to search for a counterfactual that is closest to the target factual image.
Finally, to further improve interpretability, a difference map is generated to highlight the regions of the image that were modified to create the counterfactual image.
Overall, our contributions are two-fold:
\begin{itemize}
	\item[--] We introduce DocVCE, a diffusion-based visual counterfactual explanation framework for document images that is capable of generating actionable counterfactual explanations for the task of document image classification. To the best of the authors' knowledge, this is the first work to explore diffusion-based counterfactual explanations for document image analysis.
	\item[--] We present a comprehensive qualitative and quantitative evaluation of our proposed approach on a diverse set of document benchmark datasets: \rvlcdip{}, Tobacco3482, and \doclaynet{}. Furthermore, using several well-established quantitative metrics, we demonstrate that our approach outperforms existing state-of-the-art approach LDCE~\cite{farid2023latentdiffusioncounterfactualexplanations} in terms of validity, closeness, and realism.
\end{itemize}

\section{Related Work}
\label{sec:related-work}
\subsection{eXplainable AI (XAI)}
The field of eXplainable Artificial Intelligence (XAI) has emerged as a prominent area of research~\cite{intro-survey-1,intro-survey-2}, with numerous methods developed for explaining the decisions of machine learning models~\cite{integrated-gradients,shap,lime,tcav,jeanneret2022diffusion}. These methods can be broadly categorized into three primary types: feature-attribution methods~\cite{saliency,input-x-gradient,integrated-gradients,shap,lime}, concept-based explanations~\cite{tcav}, and counterfactual explanations~\cite{lang2021explaining,jeanneret2022diffusion}.

\subsubsection{Feature-attribution methods}
Feature-attribution methods explain a model's decision-making by assigning importance to the input (or intermediate) features and are particularly valued for their post-hoc applicability. These methods can be further divided into two main types: gradient-based attribution~\cite{saliency,grad-cam,integrated-gradients,deeplift} and perturbation-based attribution~\cite{occlusion,lime,shap}. Gradient-based methods~\cite{saliency,grad-cam,integrated-gradients,deeplift} estimate the importance of features by computing the gradient of the model's prediction with respect to the input. On the other hand, perturbation-based methods~\cite{occlusion,lime,shap} perturb or ablate the input (or intermediate) features to assess their relevance for the model's decision, either by directly measuring the model's confidence scores~\cite{occlusion} or through surrogate models, such as in LIME~\cite{lime} or SHAP~\cite{shap}.

\subsubsection{Concept-based explanations}
In contrast to feature-attribution methods, concept-based approaches rely on human-comprehensible concepts to provide global explanations for a machine learning model.
A notable concept-based method is TCAV~\cite{tcav}, which utilizes directional derivatives (towards concepts) to assess the sensitivity of a model's prediction to a given concept---such as colors and textures in natural images---thus quantifying their importance.

\subsubsection{Counterfactual explanations}
Counterfactual-based explanation methods have gained considerable traction in recent years, primarily due to their ability to produce realistic, actionable explanations~\cite{Augustin2022Diffusion,jeanneret2022diffusion,farid2023latentdiffusioncounterfactualexplanations}, a property that closely aligns with the "right to explain" requirement under the GDPR regulatory framework~\cite{GDPR2016a}.
These methods typically operate on the principle of finding the minimal human-comprehensible change in the input that alters the model's decision to a  different outcome.
A popular choice to generate such minimal changes is to utilize generative models such as the generative adversarial networks (GANs)~\cite{goodfellow_generative_2014} or the diffusion models~\cite{ddpm}.
For instance, Jeanneret~\etal~\cite{jeanneret2022diffusion} proposed classifier-guided diffusion to generate visual counterfactual explanations for natural images.
Similarly, Augustin~\etal~\cite{Augustin2022Diffusion} recently introduced diffusion-based counterfactual generation that uses cone projection of robust classifier gradients over target classifier gradients to avoid generating adversarial examples~\cite{szegedy2014intriguingpropertiesneuralnetworks}.
More recently, Farid~\etal~\cite{farid2023latentdiffusioncounterfactualexplanations} extended these ideas to latent diffusion models~\cite{rombach2022highresolutionimagesynthesislatent} with classifier-free guidance~\cite{ho2022classifierfreediffusionguidance} for generating high-resolution counterfactual explanations and proposed a novel gradient consensus mechanism to filter out the adversarial gradients during counterfactual generation.

\subsection{eXplainable AI (XAI) in Document Image Analysis}
XAI has gained significant attention in recent years within the domain of document image analysis~\cite{Saifullah2023-docxai,Saifullah2022-docxclassifier,Brini2022,pouliquen2024weaklysupervisedtraininghologram}.
For instance, in a recent comprehensive benchmark study, Saifullah~\etal~\cite{Saifullah2023-docxai} applied several state-of-the-art feature-attribution methods to the task of document image classification and presented valuable insights into the fragility of deep learning models and their tendency to readily learn spurious correlations from the document data.
In another recent work, Saifullah~\etal~\cite{Saifullah2022-docxclassifier} explored inherent interpretability for document image classification by augmenting existing models with attention-based explainability components. Brini~\etal~\cite{Brini2022} proposed an end-to-end explainability framework for the task of document layout analysis that uses gradient-based methods to generate the feature-importance maps for target labels.
Similarly, Saifullah~\etal~\cite{docxplain} proposed DocXplain, a model-agnostic perturbation-based explainability framework specifically designed for document image classification.
Most recently, Pouliquen~\etal~\cite{pouliquen2024weaklysupervisedtraininghologram} also explored the applicability of gradient-based feature-attribution methods for hologram verification in identity documents.

\section{Background}
\subsection{Diffusion Models}
\label{sec:diffusion-models}
Diffusion models~\cite{ddpm,ho2022classifierfreediffusionguidance} typically operate in two steps: forward diffusion  and reverse diffusion. For any data point sampled from the real data distribution $x_0\sim q(x_0)$, the forward diffusion process defines a fixed Markov chain $\{x_1,\dots,x_T\}$ by progressively adding Gaussian noise $\epsilon\sim\mathcal{N}(0, I)$ to the data point over $T$ timesteps:
\begin{align}
    q(x_t|x_{t-1}) &= \mathcal{N}(x_t; \sqrt{\beta_t}x_{t-1}, (1-\beta_t)\mathbf{I})
\end{align}
where $\{\beta_1\dots \beta_T\}$ is a fixed variance schedule that defines the amount of noise introduced at each timestep and is chosen such that $q(x_T)\sim\mathcal{N}(0, I)$. The forward process allows sampling $x_t$ at any given timestep $t$ in closed form:
\begin{align}
    q(x_t|x_0) &= \mathcal{N}(x_t; \sqrt{\bar{\alpha}_t}x_{t-1}, (1-\bar{\alpha}_t)\mathbf{I})\\
    x_t &= \sqrt{\bar{\alpha}_t}x_{0} + \sqrt{1-\bar{\alpha}_t}\epsilon,\quad \epsilon\sim\mathcal{N}(0, I)
    \label{eq:forward-step}
\end{align}
where $\alpha_t=1 - \beta_t$ and $\bar{\alpha}_t=\Pi_{k=1}^t\alpha_k$.
The reverse diffusion process typically involves learning to undo the forward diffusion process by approximating the reverse transitions $q(x_{t-1}|x_{t})$ as $p_\theta(x_{t-1}|x_t)$:
\begin{align}
    p_\theta(x_{t-1}|x_t) &:=\mathcal{N}(x_{t-1}; \mu_\theta(x_t, t), \Sigma_\theta(x_t, t))
    \label{eq:reverse-step}
\end{align}
In practice, a neural network is trained to predict the noise $\epsilon_\theta(x_t,t)$ added in each timestep and then obtain the mean $\mu_\theta(x_t, t)$ in Eq.~\ref{eq:reverse-step} as follows:
\begin{equation}
    \mu_\theta(x_t, t) = \frac{1}{\sqrt{\alpha_t}}(x_t - \frac{\beta_t}{\sqrt{1-\bar{\alpha}}}\epsilon_\theta(x_t,t))
\end{equation}
To generate new samples from the distribution $p_\theta(x_0)$, one begins with $x_T \sim \mathcal{N}(0, I)$ and iteratively applies Eq.~\ref{eq:reverse-step} to progressively denoise the noisy sample $x_T$ into a clean sample $x_0$. The model is typically trained with the following objective:
\begin{equation}
    \label{eq:loss}
    \mathbb{E}_{t \sim [1,T],\epsilon\sim\mathcal{N}(0, I),x_{0}\sim q(x_{0})}||\epsilon - \epsilon_\theta(x_t, t)||^2
\end{equation}
It is important to note that the the loss objective in Eq.~\ref{eq:loss} is equivalent to that of the denoising score matching-based models~\cite{song2020generativemodelingestimatinggradients} with the predicted noise $\epsilon_\theta(x_t,t)$ being an estimate of the score function: $\epsilon_\theta \propto \nabla_{x_t}\log p_\theta (x_t, t)$.

\subsection{Classifier Guidance}
\label{sec:classifier-guidance}
The above formulation describes the reverse diffusion process $p_\theta(x_{t-1}|x_t)$ for the unconditional case.
Dhariwal~\etal~\cite{dhariwal2021diffusionmodelsbeatgans} showed that one may guide the reverse diffusion process with an external noise-aware classifier $p_\phi(y|.,.):\mathcal{R}\times\{1,\dots,T\}$ by additively perturbing the predicted mean $\mu_\theta(x_t, t)$ as follows:
\begin{align}
    p_{\theta,\phi}(x_{t-1}|x_t,y) &= \mathcal{N}(\tilde{\mu}_{t}, \Sigma_\theta(x_t,t) )\\
    \tilde{\mu}_t&=\mu_\theta(x_t, t) + s\Sigma_\theta(x_t,t)\nabla_{x_t} \log p_\phi (y|x_t,t)
    \label{eq:classifier-guidance}
\end{align}
where $\nabla_{x_t} \log p_\phi (y|x_t,t)$ is the classifier gradient with respect to a particular target class $y$ at each timestep $t$, and $s$ is a hyperparameter that controls the strength of the guidance.
\subsection{Classifier-free Guidance}
In contrast to classifier-guidance, classifier-free guidance~\cite{ho2022classifierfreediffusionguidance} uses an implicit classifier $p(y|x_t,t)$, trained together with the diffusion model, to guide the reverse diffusion process. Formally, by applying Bayes' rule, we have:
\begin{equation}
    \nabla_{x_t} \log p (y|x_t,t) = \nabla_{x_t} \log p (x_t|y,t) - \nabla_{x_t} \log p (x_t,t)
    \label{eq:cfg}
\end{equation}
Using the relationship between the score-based models and diffusion models as shown by Ho~\etal~\cite{ho2022classifierfreediffusionguidance}, one can rewrite the above equation in the form of the conditional and unconditional predicted noise scores:
\begin{equation}
    \nabla_{x_t} \log p (y|x_t,t) = - \frac{1}{\sqrt{1-\bar{\alpha}}} (\epsilon_\theta(x_t,t,y) - \epsilon_\theta(x_t,t))
    \label{eq:cfg-e}
\end{equation}
where $\epsilon_\theta(x_t,t,y)$ and $\epsilon_\theta(x_t,t)$ denote the noise predicted by the model for the class-conditional and unconditional cases, respectively, and $\nabla_{x_t} \log p (y|x_t,t)$ represents the classifier score that can be used to guide the generation process towards the target class $y$.
\begin{figure}[b!]
    \includegraphics[width=\textwidth]{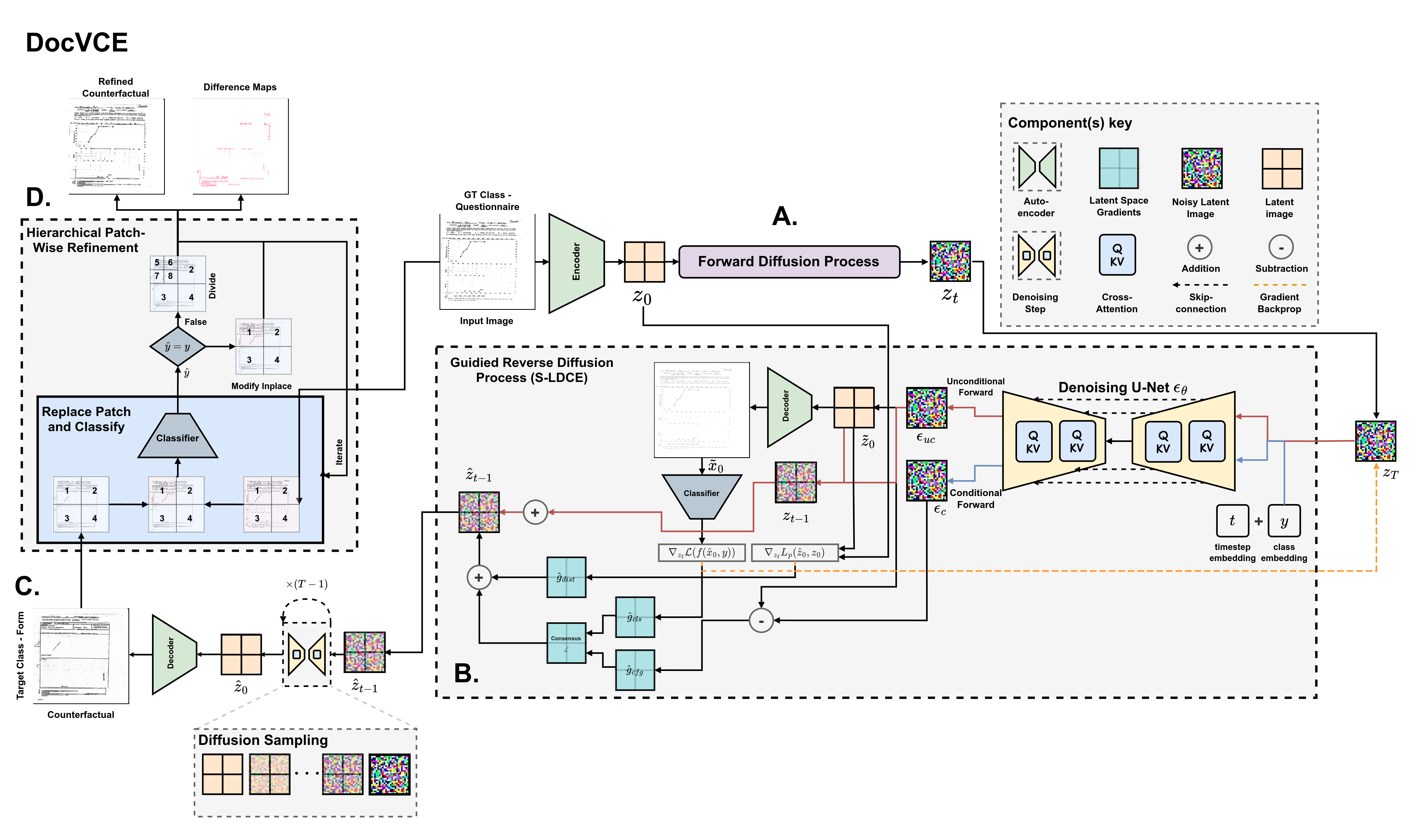}
    \caption{Overview of the proposed approach, DocVCE.
        As shown, the input image is first converted into its latent representation $z_0$ and projected to a starting noisy latent representation $z_t$ (A). Then, the noise is progressively removed by the diffusion model in each sampling step (B), guided by classifier and distance gradients, to produce the base counterfactual image (C). Finally, the base counterfactual image is fed into the hierarchical patch-wise refinement module to generate a refined counterfactual and produce a difference map (D).
     }
    \label{fig:approach}
\end{figure}
\section{DocVCE: Diffusion-based Visual Counterfactual Explanations for Document Image Classification}
In this section, we present the details of our proposed approach for generating visual counterfactual explanations for document image classification, an overview of which is provided in Fig.~\ref{fig:approach}. Given an input factual document image $x_F \in \mathbb{R}^{C\times H \times W}$, and a document classifier $p(y|x)$ that outputs the probability of a target class $y$ given $x$, a visual counterfactual explanation (VCE) can be defined as the smallest semantically meaningful and human-comprehensible change to $x_\text{F}$ that alters the prediction of the classifier $p(y|x)$ from the initial factual prediction $y = \arg \max p(y|x_F)$ to a desired target class $\tilde{y} \neq y$.
Formally, similar to the case of natural images~\cite{Augustin2022Diffusion,jeanneret2022diffusion,farid2023latentdiffusioncounterfactualexplanations}, a visual counterfactual explanation (VCE) for document images must satisfy the following properties: realism, closeness, and validity. Realism requires the the counterfactual image $x_{\text{CF}}$ should remain close to the real data manifold, making it realistic from a human perspective. Closeness means that the distance between the counterfactual image $x_{\text{CF}}$ and the factual image $x_\text{F}$ defined by some distance metric $d(x_F,x_{\text{CF}})$ must be minimal with only the most necessary semantic modifications made to the factual image $x_\text{F}$. Finally, validity requires that the classifier should predict the counterfactual image $x_{\text{CF}}$ as the desired target class $\tilde{y}$ with high probability.

Inspired by previous works~\cite{Augustin2022Diffusion,farid2023latentdiffusioncounterfactualexplanations}, we propose to generate visual counterfactual explanations for document image classifiers by guiding the reverse diffusion process of a diffusion model with classifier gradients as discussed in Section~\ref{sec:classifier-guidance}. However, given the relatively high resolution of images ($\sim 256 \times 256$) used in our task, utilizing standard diffusion models for data generation can be extremely costly. Therefore, we instead use latent diffusion models (LDMs)~\cite{rombach2022highresolutionimagesynthesislatent} and perform classifier-guided sampling in the lower-dimensional latent space of an autoencoder network $z \in \mathbb{R}^{C_z \times \frac{H}{f} \times \frac{W}{f}} = \mathcal{D}(\mathcal{E}(x))$, where $f$ is the downscaling factor of the autoencoder.
However, note that Eq.~\ref{eq:classifier-guidance} cannot be directly applied to our case, as it uses the gradients of a noise-aware classifier, which is trained on noisy images for all timesteps. Since a standard classifier, for which we need to generate a counterfactual explanation, is typically trained on clean images, its gradients can only be computed for a clean sample $x_0$ at each timestep. To this end, we rewrite Eq.~\ref{eq:forward-step} to derive a mapping for the predicted clean image $x_0$ at each timestep $t$ of the reverse sampling process:
\begin{equation}
    \tilde{x}_{0}(z_t, t) = \mathcal{D}(\frac{\tilde{z}_{0}(z_t, t)}{\eta}) = \mathcal{D}(\frac{z_t - \sqrt{1-\bar{\alpha}_t}\epsilon_\theta(z_t,t)}{\eta\sqrt{\bar{\alpha}_t}})
\end{equation}
where $\eta$ is the latent space scale factor generally employed in latent diffusion~\cite{rombach2022highresolutionimagesynthesislatent} to rescale the latent vectors $z$ to unit variance.
Notice that for latent diffusion, the sampling is done in the latent space $z \in \mathbb{R}^{C_z \times \frac{H}{f} \times \frac{W}{f}}$ and therefore the predicted clean latent image must be decoded into the pixel space $\mathbb{R}^{C\times H \times W}$ before it can be fed to the classifier $p(y|x)$.
With this formulation, the classifier gradients in Eq.~\ref{eq:classifier-guidance} can now be rewritten as $\nabla_{z_t} \log p_\phi (\tilde{y}|\tilde{x}_{0}(z_t, t),t)$\footnote{Note that the gradient $\nabla_{z_t}$ is with respect to the noisy latent space vector $z_t$ at timestep $t$, and therefore the gradients must be fully back-propagated through $\tilde{x}_{0}(z_t, t)$ instead of using an estimation as done in DiME~\cite{jeanneret2022diffusion}. } and can be used to guide the sampling process for the desired class $\tilde{y}$. Note that since diffusion generates new samples iteratively starting from pure noise, without any sort of distance function to guide the process towards the original factual image $x_\text{F}$, we cannot achieve the closeness property. To resolve this, we can add an additional guidance signal that reduces the distance between the factual image $x_\text{F}$ and the generated counterfactual image $x_{\text{CF}}$.
To this end, we minimize the standard squared L2-norm $d(\tilde{z}_0(z_t, t), z_F) = \| z_F - \tilde{z}_0(z_t, t) \|_2^2$ between the latent-space representations of the factual image $z_F$ and the predicted clean image $\tilde{z}_{0}(z_t, t)$ at each timestep, with the final updated guidance function defined as:
\begin{equation}
    \tilde{\mu}_t=\mu_\theta(z_t, t) + s\Sigma_\theta(z_t,t)\nabla_{z_t}[\lambda_c \log p_\phi (\tilde{y}|\tilde{x}_{0}(z_t, t),t) - \lambda_d d(\tilde{z}_0(z_t, t), z_F)]
\end{equation}
In addition, we use the gradient reparameterization proposed by Augustin~\etal~\cite{Augustin2022Diffusion} to redefine the guidance update rule in terms of normalized gradients which greatly simplifies the tuning of the guidance weights $\lambda_c$ and $\lambda_d$:
\begin{align}
    \label{eq:finalguidance}
    \tilde{\mu}_t&=\mu_\theta(z_t, t) + s\Sigma_\theta(z_t,t) \|\mu_\theta(z_t, t)\|_2 g\\
    g&=[\lambda_c \frac{\nabla_{z_t}\log p_\phi (\tilde{y}|\tilde{x}_{0}(z_t, t),t)}{\nabla_{z_t}\|\log p_\phi (\tilde{y}|\tilde{x}_{0}(z_t, t),t)\|_2} - \lambda_d\frac{\nabla_{z_t} d(\tilde{z}_0(z_t, t), z_F)}{\|\nabla_{z_t} d(\tilde{z}_0(z_t, t), z_F)\|_2}]
\end{align}
One major problem with direct diffusion-based classifier guidance is that it generally produces  adversarial examples~\cite{Augustin2022Diffusion,farid2023latentdiffusioncounterfactualexplanations}. We also noticed this for document images during our preliminary experiments. To address this, we employ the gradient consensus mechanism in combination with the gradients of the implicit classifier (of classifier-free guidance) proposed by Farid~\etal~\cite{farid2023latentdiffusioncounterfactualexplanations}. In particular, instead of directly using the classifier gradients, we filter out the adversarial gradients of the target classifier $\nabla_{z_t} \log p_\phi (\tilde{y}|\tilde{x}_{0}(z_t, t),t)$ by computing its patch-wise\footnote{While Farid~\etal~\cite{farid2023latentdiffusioncounterfactualexplanations} propose a patch-wise gradient threshold mechanism, in practice they use a patch size of $1 \times 1$, which we also found workable for our case.} angle with the approximate gradients of the implicit classifier $\nabla_{z_t}\log p_\theta (y|z_t,t)$ (see Eq.~\ref{eq:cfg}) and setting the gradients above a given angular threshold to zero.

Overall, with the final guidance update in Eq.~\ref{eq:finalguidance} combined with the gradient consensus mechanism, our approach closely resembles the previously introduced state-of-the-art LDCE~\cite{farid2023latentdiffusioncounterfactualexplanations} method.
In particular, our approach can be viewed as a slightly different formulation of the LDCE~\cite{farid2023latentdiffusioncounterfactualexplanations} method, which uses stochastic guided sampling~\cite{dhariwal2021diffusionmodelsbeatgans} instead of deterministic guided sampling~\cite{dhariwal2021diffusionmodelsbeatgans} for counterfactual generation. Accordingly, we refer to this approach as S-LDCE.

\subsection{Hierarchical Patch-wise Counterfactual Refinement}
Unlike natural image classification, document image classification often depends on global semantic features like structural layout, font styles, and text formatting.
As a result, when gradients of such classifiers are used to guide the generation of a counterfactual image $x_{\text{CF}}$, they often lead to altering the entire image instead of making modifications to specific localized regions.
While such counterfactuals already provide valuable insights into the classifier's properties, they can be difficult to interpret due to the numerous modifications made to the image. Furthermore, they may not achieve sufficient closeness to the factual image $x_{\text{F}}$.
To address this, we propose a hierarchical patch-wise refinement (HPR) strategy that iteratively refines the counterfactual image $x_{\text{CF}}$ by converting regions that are unimportant for the desired target label $\tilde{y}$ back to the original factual image $x_{\text{F}}$.

Formally, given an initial counterfactual image $x_{\text{CF}}$ predicted as the desired counterfactual label $\tilde{y}$, we recursively divide the image into a patch-grid where the initial patch size is $\frac{H}{2} \times \frac{W}{2}$. For each patch, we sequentially replace the corresponding region of the counterfactual image $x_{\text{CF}}$ with the target factual image $x_{\text{F}}$, which results in a perturbed counterfactual image $\tilde{x}_{\text{CF}}$. After each replacement, we compute the predicted class label $\hat{y}=\arg\max(p(y|\tilde{x}_{\text{CF}}))$, and compare it with the desired counterfactual label $\tilde{y}$.
If the prediction remains unchanged, i.e., $\hat{y} = \tilde{y}$, and its probability remains within a given threshold $\delta$, it means that this patch may not be as important to the desired counterfactual class, and therefore, the patch is permanently replaced with the corresponding region from $x_{\text{F}}$. However, if the prediction changes, we further subdivide the patch into another patch-grid of size $\frac{p}{2} \times \frac{p}{2}$ and append these to the list of patches to process.
This process is repeated recursively until a minimum patch size is reached, which we set to $16 \times 16$ in this work.
Note that, while the HPR algorithm requires sequential computation of classifier probabilities (which may be costly), this computation can be efficiently batched for processing multiple images together.
Finally, since the HPR component is specifically designed for sparse, semi-structured document-like images, which are separated into a background and foreground, it may not perform well when applied to natural images. As a result, we refer to the combination of the proposed S-LDCE and HPR approaches as DocVCE.
For the complete pseudo-code implementation of the proposed DocVCE algorithm, see Appendix~\ref{app:pseudocodes}.

\section{Experiments and Results}
\subsection{Datasets}
We evaluate our approach on 3 diverse document image classification datasets: (1) \rvlcdip{}, (2) Tobacco3482\footnote{https://www.kaggle.com/datasets/patrickaudriaz/tobacco3482jpg}, and (3) DocLayNet~\cite{doclaynet}. \rvlcdip{} is a large-scale document benchmark dataset that has been extensively used in previous works for the document classification task~\cite{afzal2017-doc-class-2,ferrando2020-doc-class-4,Saifullah2022-docxclassifier,layoutlmv3}. The dataset consists of 400,000 business document images, categorized into 16 distinct document types, and features highly diverse document layouts and styles. It is split into training, validation, and testing sets of sizes 320,000, 40,000, and 40,000, respectively. Tobacco3482, in contrast, is a relatively small-scale dataset consisting of only 3,482 business documents spread across 10 different document classes. It also exhibits a highly imbalanced class distribution. DocLayNet~\cite{doclaynet} is a recently introduced dataset containing modern documents categorized into 6 distinct document types. It is split into training, validation, and testing sets of sizes 69,375, 6,489, and 4,999, respectively.
\subsection{Implementation Details}
\subsubsection{Autoencoder pretraining}
Since the target classifier, the gradients of which are used to guide the sampling process, is trained on the real data distribution, it is essential to ensure a sufficiently high reconstruction quality from the autoencoder to generate viable counterfactual explanations.
While existing pretrained autoencoders provided by Rombach~\etal~\cite{rombach2022highresolutionimagesynthesislatent} offer adequate reconstruction accuracy on medium-resolution ($256 \times 256$) document images, we observed that, on gray-scale document images, the reconstruction sometimes introduced RGB-colored artifacts.
To avoid such artifacts, we fine-tune a KL-reg autoencoder variant proposed by Rombach~\etal~\cite{rombach2022highresolutionimagesynthesislatent} on 11M document images from the large-scale IIT-CDIP Test Collection 1.0~\cite{iitcdip}, using a combination of adversarial and perceptual losses.
In this work, we utilize an autoencoder with a downscaling factor of $f=4$, which compresses the input document images of dimensions $3 \times 256 \times 256$ to a latent space of dimensions $3 \times 64 \times 64$.
\subsubsection{Model architecture}
For noise prediction, we use a UNet-based diffusion model proposed by Ho~\etal~\cite{ho2022classifierfreediffusionguidance} that consists of timestep-conditioned residual blocks~\cite{resnet}, along with self-attention applied to the last two feature map resolutions of $16 \times 16$ and $8 \times 8$.
For the diffusion process, we use a linear noise scheduler with noise ranging from $\beta_1 = 10^{-4}$ to $\beta_T = 0.02$ over a total of $T=1000$ timesteps.
We train the diffusion model with class-conditioning~\cite{dhariwal2021diffusionmodelsbeatgans}, and to enable classifier-free guidance~\cite{ho2022classifierfreediffusionguidance} during the sampling phase, we replace the class labels with a null label with a probability of 0.1 during training.
\subsubsection{Counterfactual sampling}
For sampling, we apply timestep re-spacing to divide the total timestep range (T=1000) into 200 inference steps.
Note that while one could start from pure noise $\epsilon\sim\mathcal{N}(0, I)$, and use the guidance mechanism proposed in Eq.~\ref{eq:finalguidance} to generate a counterfactual, in practice, this approach makes it extremely challenging for the model to produce a counterfactual that remains close to the original factual image, $x_\text{F}$.
Therefore, following previous works~\cite{Augustin2022Diffusion,farid2023latentdiffusioncounterfactualexplanations,jeanneret2022diffusion}, we instead project $z_F$ into an intermediate noisy representation, $z_{t'}$ which is then used as the  starting point for the reverse diffusion process.
In this work, we experiment with multiple settings for the starting timestep $t'\in\{60,80,100\}$. In addition, we explore various values for the guidance scale $s\in\{1.5,2.0,3.0\}$, as well as different configurations of the the guidance weights $(\lambda_c,\lambda_d)=\{(0.8,0.2),(0.7,0.3),(0.6,0.4)\}$.
For consensus-based filtration, we set the angle threshold to a fixed value of $45.0^\circ$.

\subsection{Evaluation protocol}
For a rigorous qualitative and quantitative assessment of our approach, we generate visual counterfactual explanations for a total of 10,000 random samples from the \rvlcdip{} test set, 2,500 random samples from the \doclaynet{} test set, and 1,000 random samples from the full Tobacco3482 dataset.
Following the work of Farid~\etal~\cite{farid2023latentdiffusioncounterfactualexplanations}, the target counterfactual label $\tilde{y}$ is randomly sampled as one of the top-5 (top-3 in case of \doclaynet{}) closest classes, where the closeness between classes is determined based on the instance-wise cosine similarity of SimSiam~\cite{chen2020simsiam} features.
For this purpose, we train a ResNet-50~\cite{resnet} model with SimSiam~\cite{chen2020simsiam} representation-learning strategy on all three datasets.

We use several quantitative evaluation metrics available in the literature~\cite{Augustin2022Diffusion,jeanneret2022diffusion,farid2023latentdiffusioncounterfactualexplanations} for evaluating the validity, realism, and closeness of the generated counterfactuals.
In particular, we utilize the flip ratio (FR) and mean confidence scores (Mean Conf.) to evaluate validity, as both metrics measure the overall accuracy of the counterfactual generation process in reaching the desired target label $\tilde{y}$.
To measure closeness, we compute the $l_1$ norm, $l_2$ norm, and LPIPS loss between the counterfactual image $x_\text{CF}$ and the factual image $x_\text{F}$.
To assess realism, we use the \text{Fr\'{e}chet Inception Distance (FID)} and sFID~\cite{jeanneret2022diffusion} scores between the factual and counterfactual datasets. We perform all quantitative evaluations on 3 different document image classifiers: ResNet-50~\cite{resnet}, ConvNeXt-B~\cite{convnext}, and DiT-B~\cite{doc-vit}.
\begin{figure}[t!]
\centering
\begin{subfigure}{0.2\textwidth}
    \centering
    \includegraphics[width=\textwidth]{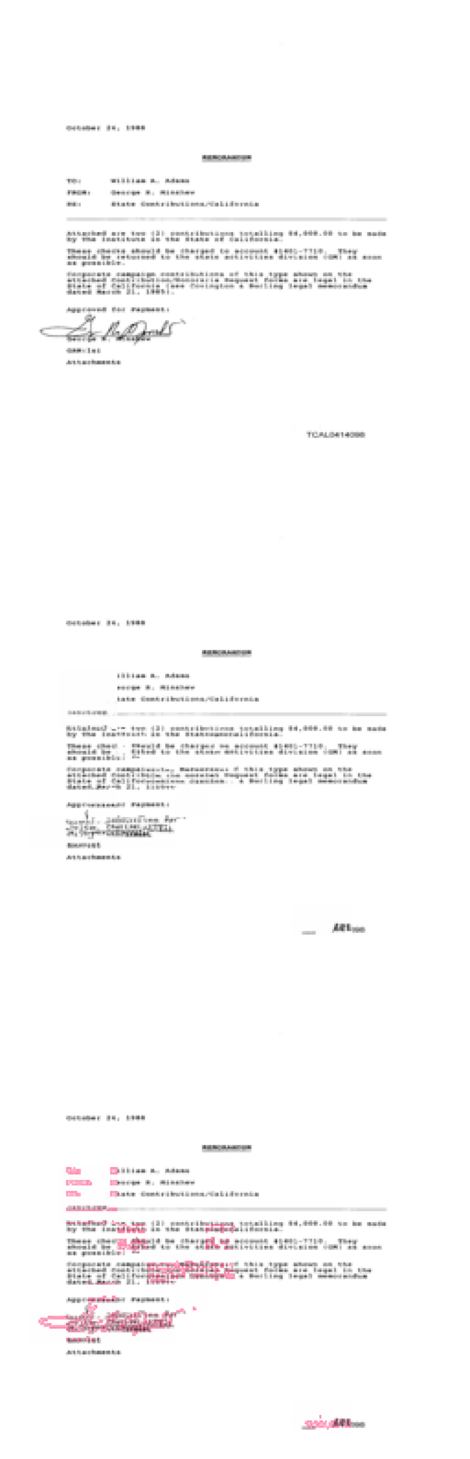}
    \caption{memorandum $\rightarrow$ resume}
    \label{fig:qual1}
\end{subfigure}
\begin{subfigure}{0.2\textwidth}
    \centering
    \includegraphics[width=\textwidth]{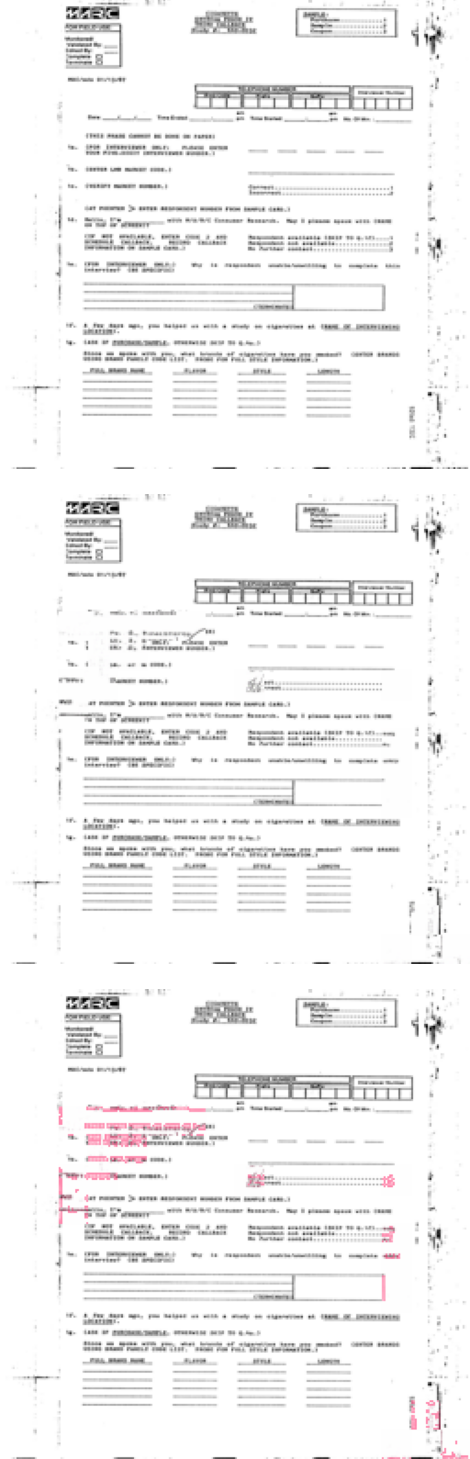}
    \caption{questionnaire $\rightarrow$ memorandum}
    \label{fig:qual2}
\end{subfigure}
\begin{subfigure}{0.2\textwidth}
    \centering
    \includegraphics[width=\textwidth]{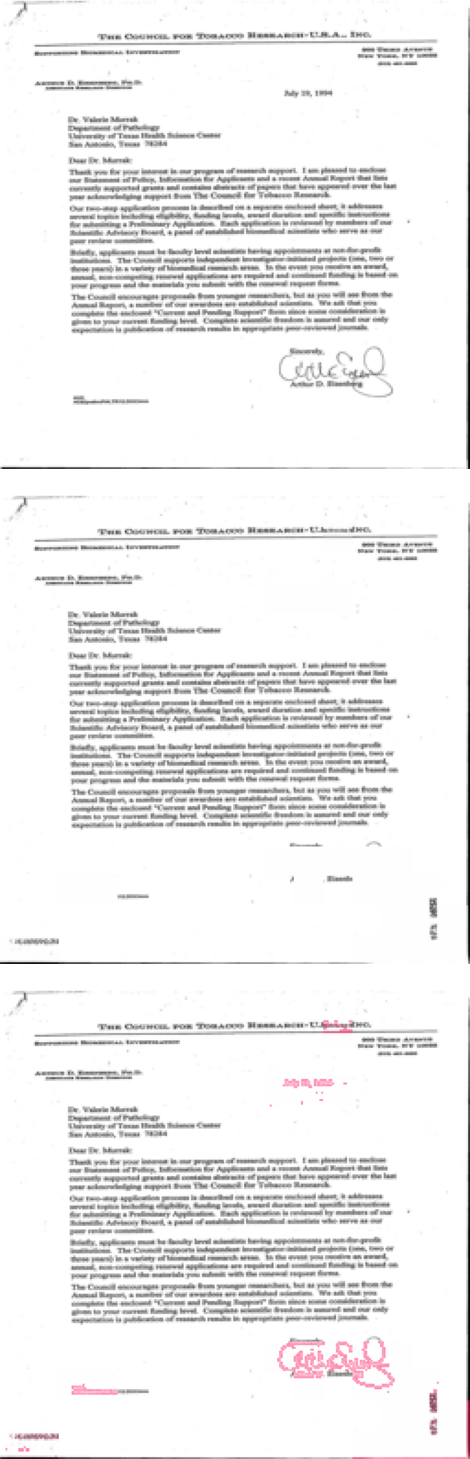}
    \caption{letter $\rightarrow$\\email}
    \label{fig:qual3}
\end{subfigure}
\begin{subfigure}{0.2\textwidth}
    \centering
    \includegraphics[width=\textwidth]{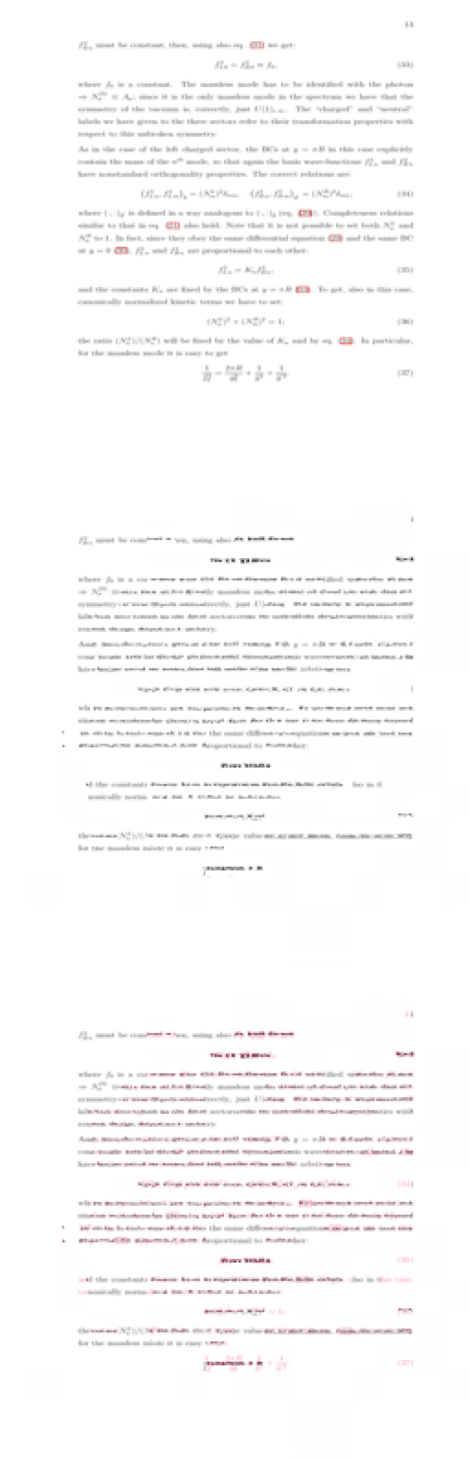}
    \caption{sci. article $\rightarrow$ gov. tender}
    \label{fig:qual4}
\end{subfigure}
\caption{Counterfactual explanations generated using the proposed DocVCE framework for different datasets and classifiers. For each sample, we display the original document image (top), the corresponding counterfactual explanation (middle), and the difference map (bottom), which highlights the regions of the image that were modified to generate the counterfactual.}
\label{fig:qualitative-results-1}
\end{figure}
\begin{figure}[t!]
\centering
\begin{subfigure}{0.9\textwidth}
    \centering
    \includegraphics[width=0.9\textwidth]{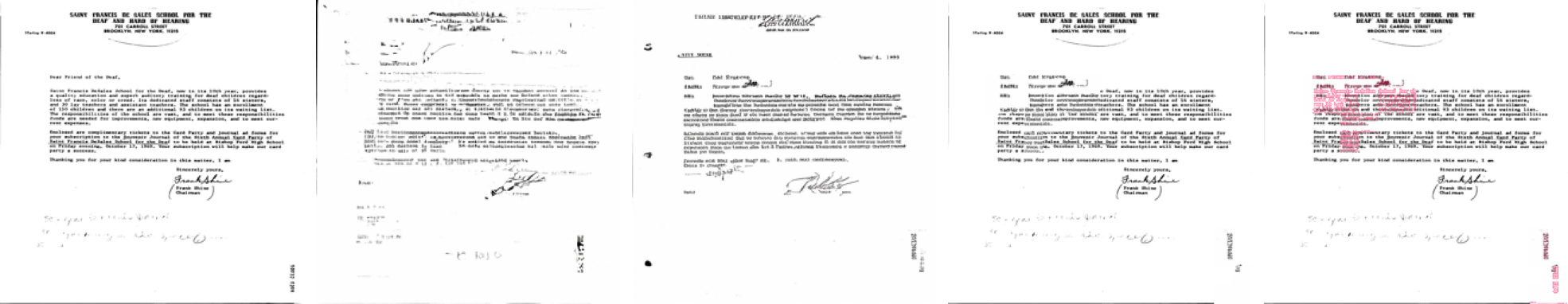}
    \caption{letter$\rightarrow$memo}
    \label{fig:qual21}
\end{subfigure}
\begin{subfigure}{0.9\textwidth}
    \centering
    \includegraphics[width=0.9\textwidth]{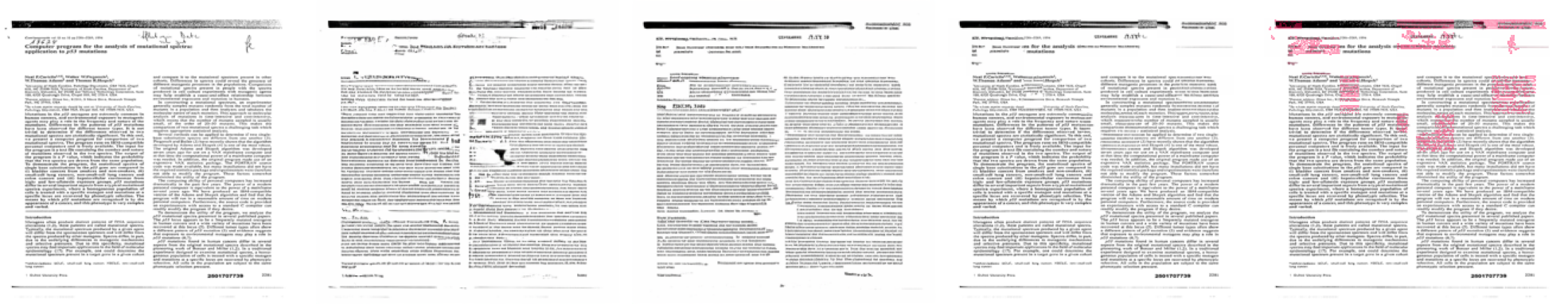}
    \caption{sci. pub.$\rightarrow$email}
    \label{fig:qual22}
\end{subfigure}
\begin{subfigure}{0.9\textwidth}
    \centering
    \includegraphics[width=0.9\textwidth]{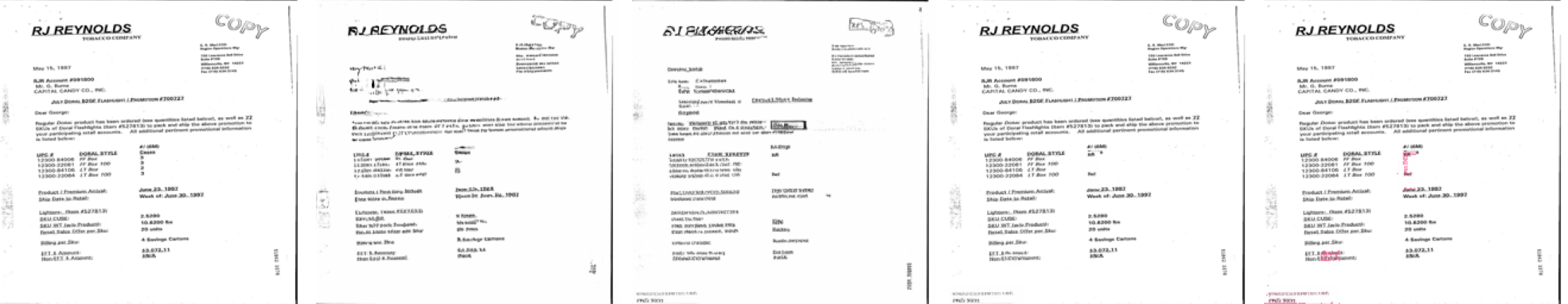}
    \caption{letter$\rightarrow$email}
    \label{fig:qual23}
\end{subfigure}
\begin{subfigure}{0.9\textwidth}
    \centering
    \includegraphics[width=0.9\textwidth]{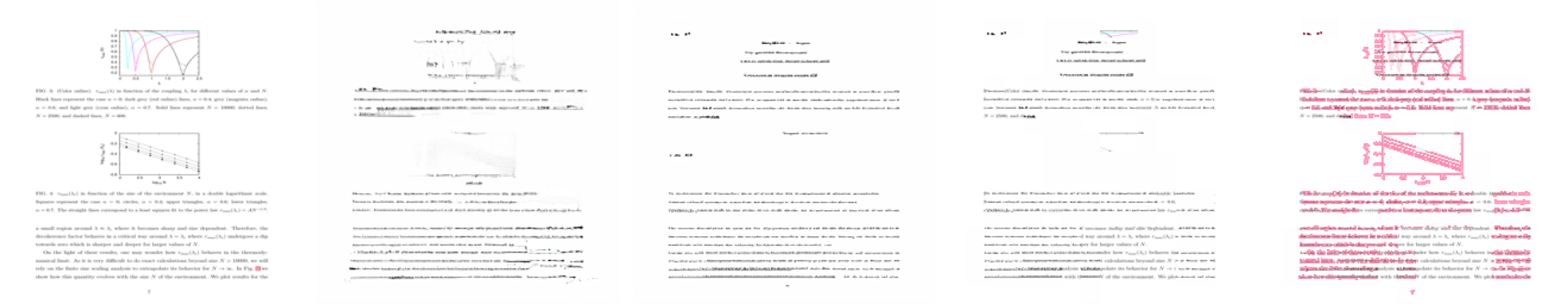}
    \caption{sci. article$\rightarrow$gov. tender}
    \label{fig:qual24}
\end{subfigure}
\caption{Qualitative comparison between the proposed methods, S-LDCE and DocVCE, and the existing state-of-the-art approach \ldce{}. Left to right: original image, LDCE~\cite{farid2023latentdiffusioncounterfactualexplanations}, S-LDCE (ours), DocVCE (ours), DocVCE difference map (ours).
We observe that \ldce{} often produces out-of-distribution artifacts in the counterfactuals, whereas S-LDCE results in excessive modifications to the image. In contrast, DocVCE is able to generate counterfactuals with minimal changes to the input image.
}
\label{fig:qualitative-results-2}
\end{figure}
\subsection{Qualitative Evaluation}
Fig.~\ref{fig:qualitative-results-1} and Fig.~\ref{fig:qualitative-results-2} present qualitative results for the proposed DocVCE approach across different dataset samples and classifiers.
We observe that DocVCE is capable of generating minimal yet realistic modifications to an image, both local (as shown in Fig.\ref{fig:qual1}, Fig.~\ref{fig:qual2}, Fig.~\ref{fig:qual3}) and global (see Fig.~\ref{fig:qual4}), to generate visual counterfactual explanations across a diverse range of document images.
We can identify key patterns for different classes by analyzing the counterfactuals. For instance, in Fig.\ref{fig:qual1} and Fig.\ref{fig:qual2}, we observe that the header containing the sender and recipient names is particularly important for identifying the memorandum class, as in Fig.~\ref{fig:qual1} the header is removed to change the prediction, while in Fig.~\ref{fig:qual2}, the top left area is modified by the model to resemble a memorandum header with a list of names.
Similarly, from Fig.~\ref{fig:qual3}, the signature, date, and document ID numbers seem like important elements for identifying the letter and email classes, as the first two elements (signature and date) are removed from the image, while new document IDs with specific styles (in the bottom-left and bottom-right corners) are introduced by the model to generate the counterfactual.
Lastly, in Fig.~\ref{fig:qual4}, when converting a scientific article to the government tender type, the model pays close attention to the font styles and equations. In particular, it transforms all equations into individual paragraph headers, indicating that it considers equations as key features for classifying scientific articles. Additional qualitative results for counterfactuals generated on various datasets and models using DocVCE are provided in Appendix~\ref{app:qualitative-results}.
\begin{figure}[t]
    \centering
    \includegraphics[width=0.8\textwidth]{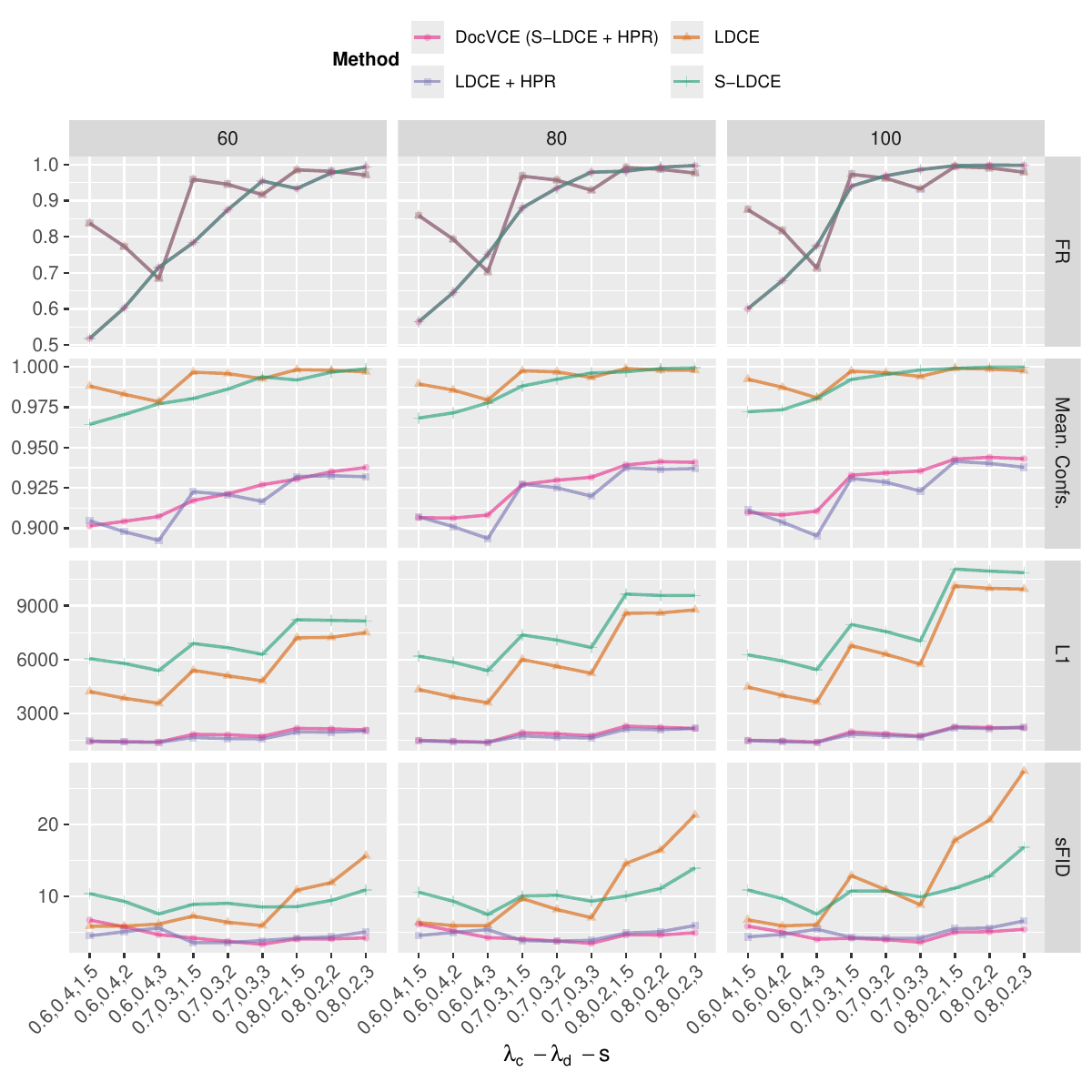}
    \caption{
        Ablation study showing the effect of various hyperparameters, such as starting timestep $t'$, classifier and distance weights $(\lambda_c,\lambda_d)$, and guidance scale $s$, on the validity, closeness, and realism of document counterfactuals. All results were computed on \rvlcdip{} dataset with ConvNeXt-B~\cite{convnext} classifier. The guidance weights $(\lambda_c,\lambda_d)=(0.7,0.3)$ produce the best results across all setups.
    }
    \label{fig:ablation}
    \vspace{-2em}
\end{figure}

\begin{table}[h]
    \centering
    \setlength{\tabcolsep}{0.5em}
    \resizebox{0.8\textwidth}{!}{
        \begin{tabular}{l|l|lccccccc}
            \toprule
            \multicolumn{3}{c}{}&\multicolumn{2}{c}{Validity}&\multicolumn{3}{c}{Closeness}&\multicolumn{2}{c}{Realism}\\
            \cmidrule(lr){4-5}\cmidrule(lr){6-8}\cmidrule(lr){9-10} \multicolumn{2}{c}{}&Method&FR ($\uparrow$) & Mean Confs. ($\uparrow$) & L1 ($\downarrow$) & L2 ($\downarrow$) & LPIPS ($\downarrow$) & FID ($\downarrow$) & sFID ($\downarrow$) \\
            \midrule
            \multirow{12}{*}{\rotatebox{90}{RVL-CDIP~\cite{harley2015-rvlcdip}}}
            &\multirow{4}{*}{\rotatebox{90}{ResNet}}
            &\ldce&81.62&97.84&6142.53&36.07&0.11&7.73&9.31\\
            &&\sldce&84.48&97.58&6723.35&39.27&0.14&8.98&9.49\\
            &&\ldcehpr&81.62&88.86&1967.13&21.32&0.04&2.27&5.29\\
            &&\sldcehpr&84.48&89.62&1791.65&21.27&0.04&2.57&4.14\\
            \cmidrule{2-10}
            &\multirow{4}{*}{\rotatebox{90}{ConvNeXt}}
            &\ldce&93.25&99.41&5746.53&34.40&0.11&8.15&8.81\\
            &&\sldce&98.60&99.81&7034.11&40.21&0.15&9.80&9.92\\
            &&\ldcehpr&93.25&92.30&1706.48&20.17&0.03&2.35&4.18\\
            &&\sldcehpr&98.60&93.54&1739.21&21.10&0.04&2.53&3.62\\
            \cmidrule{2-10}
            &\multirow{4}{*}{\rotatebox{90}{DiT}}
            &\ldce&92.28&99.35&5824.87&35.07&0.11&8.25&8.96\\
            &&\sldce&99.11&99.85&7159.22&40.99&0.15&10.04&10.33\\
            &&\ldcehpr&92.28&91.90&1848.40&21.48&0.04&2.50&4.35\\
            &&\sldcehpr&99.11&93.20&1681.18&20.99&0.04&2.17&3.58\\
            \cmidrule{1-10}
            \multirow{12}{*}{\rotatebox{90}{Tobacco3482}}
            &\multirow{4}{*}{\rotatebox{90}{ResNet}}
            &\ldce&77.50&95.28&10961.50&59.39&0.24&39.79&44.69\\
            &&\sldce&80.60&96.38&9634.62&53.54&0.23&20.22&26.70\\
            &&\ldcehpr&77.50&86.55&4143.76&36.39&0.09&13.80&23.50\\
            &&\sldcehpr&80.60&87.57&2978.63&30.48&0.08&9.66&19.76\\
            \cmidrule{2-10}
            &\multirow{4}{*}{\rotatebox{90}{ConvNeXt}}
            &\ldce&95.20&99.25&10034.16&55.63&0.26&44.75&46.91\\
            &&\sldce&97.40&99.07&10289.83&55.50&0.26&23.86&25.78\\
            &&\ldcehpr&95.20&90.59&2996.45&30.73&0.08&13.18&18.90\\
            &&\sldcehpr&97.40&90.68&2862.50&29.20&0.08&11.05&15.98\\
            \cmidrule{2-10}
            &\multirow{4}{*}{\rotatebox{90}{DiT}}
            &\ldce&94.90&99.61&9808.09&55.05&0.24&43.39&46.01\\
            &&\sldce&98.40&99.76&9969.83&54.51&0.24&22.22&24.06\\
            &&\ldcehpr&94.90&92.69&3012.48&30.73&0.07&13.50&19.22\\
            &&\sldcehpr&98.40&93.10&2624.06&27.71&0.07&10.22&15.66\\
            \cmidrule{1-10}
            \multirow{12}{*}{\rotatebox{90}{DocLayNet~\cite{doclaynet}}}
            &\multirow{4}{*}{\rotatebox{90}{ResNet}}
            &\ldce&97.16&94.76&6311.79&33.12&0.16&33.78&35.13\\
            &&\sldce&89.12&88.00&6962.76&36.58&0.17&20.60&23.76\\
            &&\ldcehpr&97.16&79.74&2386.64&20.53&0.07&8.53&12.59\\
            &&\sldcehpr&89.12&69.03&1810.51&18.71&0.06&5.42&11.57\\
            \cmidrule{2-10}
            &\multirow{4}{*}{\rotatebox{90}{ConvNeXt}}
            &\ldce&98.84&96.33&5354.85&27.59&0.15&31.42&32.86\\
            &&\sldce&96.36&94.65&6952.97&36.20&0.17&21.17&22.61\\
            &&\ldcehpr&98.84&87.25&2467.26&19.07&0.07&11.16&14.11\\
            &&\sldcehpr&96.36&85.75&2837.23&23.13&0.09&9.91&13.03\\
            \cmidrule{2-10}
            &\multirow{4}{*}{\rotatebox{90}{DiT}}
            &\ldce&98.76&96.56&5810.40&30.47&0.15&33.10&34.46\\
            &&\sldce&98.84&97.44&7353.46&37.84&0.18&21.42&22.44\\
            &&\ldcehpr&98.76&88.51&2423.97&20.22&0.08&10.37&14.40\\
            &&\sldcehpr&98.84&88.47&2895.06&23.82&0.09&7.87&11.82\\
            \bottomrule
        \end{tabular}
    }
    \caption{
        Quantitative results of our proposed approaches, S-LDCE and DocVCE, in comparison to LDCE~\cite{farid2023latentdiffusioncounterfactualexplanations} on \rvlcdip{}, Tobacco3482, and DocLayNet~\cite{doclaynet} with 3 different document classification models: ResNet-50~\cite{resnet}, ConvNeXt-B~\cite{convnext}, and DiT-B~\cite{doc-vit}. We observe that with the HPR module, both the closeness and realism of the generated counterfactuals are significantly improved, but at the cost of mean confidence scores.
    }
    \label{table:results}
    \vspace{-1em}
\end{table}
In Fig.~\ref{fig:qualitative-results-2}, we present a qualitative comparison between the proposed methods, S-LDCE and DocVCE, and the existing state-of-the-art approach \ldce{}\footnote{Note that in this work, we reproduce \ldce{} for document images, with distance gradients computed in the latent space, consistent with the S-LDCE and DocVCE implementations.}. We observe that while \ldce{} has shown promising results on natural images, its application to document images generally led to artifacts, and out-of-distribution modifications to the input images (as seen in Fig.~\ref{fig:qual21}, Fig.~\ref{fig:qual22}, and Fig.~\ref{fig:qual23}).
In contrast, S-LDCE, with its stochastic sampling (as opposed to deterministic sampling in \ldce{}), was able to generate more realistic, in-distribution counterfactuals for document images.
This suggests that introducing randomness during sampling may play a crucial role in correcting the diffusion trajectory after each noisy classifier-guidance update.
Note that S-LDCE is already capable of generating useful and realistic document counterfactuals that can provide valuable insights into a classifier's prediction.
However, we find that counterfactuals generated by S-LDCE are often difficult to interpret due to the excessive global modifications made to the document.
On the other hand, DocVCE is able to search for only the most minimal modifications that are required to flip the classifier's prediction, ultimately improving the interpretability of the explanation by both localizing the updates, and generating a difference map.
For instance, as observed in Fig.~\ref{fig:qual21}, S-LDCE updates the title, header, and the signature in the document to flip the classifier decision from letter to memo class.
However, with DocVCE, it becomes visible that these modifications are unnecessary, and that the only modification required to flip the prediction is updating the header.
This enables a user to directly map a change in prediction to a specific element, feature, or concept in the image, as opposed to S-LDCE, where it is difficult to determine which modifications had the biggest (or smallest) effect in flipping the class prediction.
Additional results for the qualitative comparison of different approaches are provided in Appendix~\ref{app:additional-qualitative-comparison}.

\subsection{Quantitative Evaluation}
In Fig.~\ref{fig:ablation}, we present the results of our ablation study, which was conducted to determine the appropriate hyperparameters for generating the explanations.
We observe that for all guidance scales, a higher classifier weight $(\lambda_c,\lambda_d)=(0.8,0.2)$, significantly degraded the L1 and sFID, whereas using a lower classifier weight $(\lambda_c,\lambda_d)=(0.6,0.4)$ resulted in severe drops in FR.
Therefore, we opted for the middle ground  $(\lambda_c,\lambda_d)=(0.7,0.3)$ for generating the samples. On the other hand, we observe that lower starting timesteps $t'=60$ work better for both closeness and realism metrics; however, they can lead to slight degradation in FR. To avoid drops in FR, for all final evaluations, we use a starting timestep $t'=100$ with a guidance scale of $s=3$.

In Table~\ref{table:results}, we provide the quantitative results of our proposed approaches and compare them to the existing LDCE~\cite{farid2023latentdiffusioncounterfactualexplanations} approach.
We find that on \rvlcdip{}, \ldce{} generally outperformed S-LDCE in terms of closeness and realism; however, it significantly underperformed in terms of FR on ConvNeXt~\cite{convnext} and DiT~\cite{doc-vit} models.
In contrast, on the \tobacco{} and \doclaynet{} datasets, S-LDCE outperformed \ldce{} by a significant margin in terms of realism, while performing similarly in closeness.
When used in combination with the proposed HPR module, the closeness and realism of both LDCE and S-LDCE improve significantly; however, this comes at the cost of slight drops in model confidence.
This shows that our proposed HPR module essentially trades off mean confidence for better closeness and realism with the real image.
This also makes it useful, as it tries to find counterfactuals that are closer to the decision boundary, allowing users to better identify spurious correlations that a model might be learning.
However, it is worth noting that all confidence drops with HPR stay under the pre-defined threshold $\delta=0.1$ and therefore the model is still sufficiently confident in its predictions on the refined counterfactuals.

\section{Conclusion}
In this work, we presented the first visual counterfactual explanation approach for document image classification, capable of generating realistic minimal changes to document images to provide actionable counterfactual explanations.
Since this is the first work exploring generative visual counterfactuals for document images, we focused solely on the task of document image classification.
However, in the future, it will be worthwhile to extend this approach to more complex document image analysis tasks, such as handwriting recognition, layout analysis, and table extraction.
In addition, our current approach can only generate counterfactual explanations for medium-sized resolutions of $256\times256$.
However, with the recent rise of OCR-free end-to-end transformers~\cite{kim2022donut} and vision-language models (VLMs)~\cite{tang2023unifyingvisiontextlayout}, which extract information directly from the visual content of super-resolution document images, exploring counterfactual image generation for super-resolution document images may be worth pursuing.
Lastly, it could be a plausible future direction to explore document inpainting for modifying only targeted regions in the image to generate the counterfactuals.
\bibliographystyle{splncs04}
\bibliography{references.bib}
\newpage
\appendix
\section{Algorithm implementations}
\label{app:pseudocodes}
In Algorithm~\ref{alg:docvce}, we present the pseudocode for the proposed DocVCE counterfactual generation strategy. As shown, the factual image is first projected into the latent space and then converted into a noisy latent image at timestep $t'$. Next, we iteratively apply classifier- and distance-based guidance during the sampling process to generate the base counterfactual image.
\begin{algorithm}[t]
    \caption{Document visual counterfactual explanations (DocVCE).}\label{alg:docvce}
    \begin{algorithmic}
        \State \textbf{Input:} Factual image $x_{\text{F}}$, desired target label $\tilde{y}$, target classifier $p_\phi(y|x)$, encoder $\mathcal{E}$, decoder $\mathcal{D}$, starting timestep $t'$, guidance scale s, classifier weight $\lambda_c$, distance weight $\lambda_d$, latent-space scale $\eta$, angle threshold $\gamma$, refinement confidence threshold $\delta$
        \State \textbf{Output:} refined counterfactual image $\tilde{x}_\text{CF}$
        \State $z_{t'} \leftarrow \sqrt{\bar{\alpha}_{t'}}\mathcal{E}(x^{F}) + \sqrt{1-\bar{\alpha}_{t'}}\epsilon,\quad \epsilon\sim\mathcal{N}(0, I)$
        \For{$t=t',\; \dots,\; 0$}
        \State $\tilde{x}_{0}(z_t, t) \leftarrow \mathcal{D}(\frac{z_t - \sqrt{1-\bar{\alpha}_t}\epsilon_\theta(z_t,t)}{\eta\sqrt{\bar{\alpha}_t}})$
        \State $\text{g\textsubscript{\texttt{cls}}},\text{g\textsubscript{\texttt{cfg}}} \leftarrow \nabla_{z_t}\log p_\phi (\tilde{y}|\tilde{x}_{0}(z_t, t),t), \frac{1}{\sqrt{1-\bar{\alpha}}} (\epsilon_\theta(x_t,t) - \epsilon_\theta(x_t,t,y))$
        \State $g\textsubscript{\texttt{cls-filtered}}, g\textsubscript{\texttt{dist}}$ $\leftarrow$ $\mathbf{0}$ if $\angle(g_\textsubscript{cls}, g_\textsubscript{\texttt{cfg}}) > \gamma$ else $g\textsubscript{cls}, \nabla_{z_t} d(\tilde{z}_0(z_t, t), z_F)$
        \State $g \leftarrow \lambda_c \frac{g\textsubscript{cls-filtered}}{\|g\textsubscript{cls-filtered}\|_2} - \lambda_d\frac{g\textsubscript{dist}}{\|g\textsubscript{dist}\|_2}$
        \State $\mu,\Sigma \leftarrow \mu_\theta(z_t, t), \Sigma_\theta(z_t, t)$
        \State $z_{t-1} \leftarrow \text{sample from } \mathcal{N}(\mu_\theta(z_t, t) + s\Sigma_\theta(z_t,t) \|\mu_\theta(z_t, t)\|_2 g), \Sigma_\theta(z_t,t))$
        \EndFor
        \State $x_\text{CF}\leftarrow \mathcal{D}(\frac{z_0}{\eta})$
        \State $\tilde{x}_\text{CF} \leftarrow \texttt{HPR} (x_\text{CF}, x_\text{F}, p\phi(y|x), \delta)$
    \end{algorithmic}
\end{algorithm}
\begin{algorithm}[t]
\caption{Hierarchical Patch-Wise Refinement (HPR)}\label{alg:hpr}
\begin{algorithmic}[1]
    \State \textbf{Input:} Base counterfactual image $x_{\text{CF}}$, original factual image $x_{\text{F}}$, target classifier $p_\phi(y|x)$, confidence difference threshold $\delta$
    \State \textbf{Output:} refined counterfactual image $\tilde{x}^{\text{CF}}$
    \State $\mathcal{S} \gets \texttt{create\_patch\_grid}(x_{\text{CF}})$ \Comment{Create patches from the initial image}
    \State $p_{\text{CF}}, y_{\text{CF}} \gets f(x_{\text{CF}}), \texttt{argmax}(f(x_{\text{CF}}))$ \Comment{Get initial model prediction}
    \While{!$\mathcal{S}$.\texttt{isEmpty()}}
    \State $p \gets \mathcal{S}\texttt{.pop()}$ \Comment{Pop the next patch from the list}
    \State $\tilde{x}_{\text{CF}} \gets \texttt{modify\_image\_patch}(x_{\text{CF}}, p)$
    \State $\tilde{p}_{\text{CF}}, \tilde{y}_{\text{CF}} \gets f(\tilde{x}_{\text{CF}}), \texttt{argmax}(f(\tilde{x}_{\text{CF}}))$ \Comment{Predicted class for updated image}
    \If{$y_{\text{CF}} = \tilde{y}_{\text{CF}}$ \textbf{and} $|p_{\text{CF}}[y_{\text{CF}}] - \tilde{p}[y_{\text{CF}}]| < \delta$}
    \State $x_{\text{CF}} \gets \tilde{x}_{\text{CF}}$ \Comment{Update image if confidence difference is below threshold}
    \Else
    \State $\mathcal{S} \gets \texttt{create\_patch\_grid}(p)$ \Comment{Split patch into smaller ones for further refinement}
    \EndIf
    \EndWhile
    \State \textbf{return} $\tilde{x}_{\text{CF}}$
\end{algorithmic}
\end{algorithm}
Finally, we apply the HPR strategy to the base counterfactual to generate the refined counterfactual. The full pseudocode for the HPR strategy is provided in Algorithm~\ref{alg:hpr}.
\section{Spurious Correlations in the RVL-CDIP and Tobacco3482 Datasets}
\label{app:spurios-corr}
\begin{figure}[H]
    \centering
    \includegraphics[width=\textwidth]{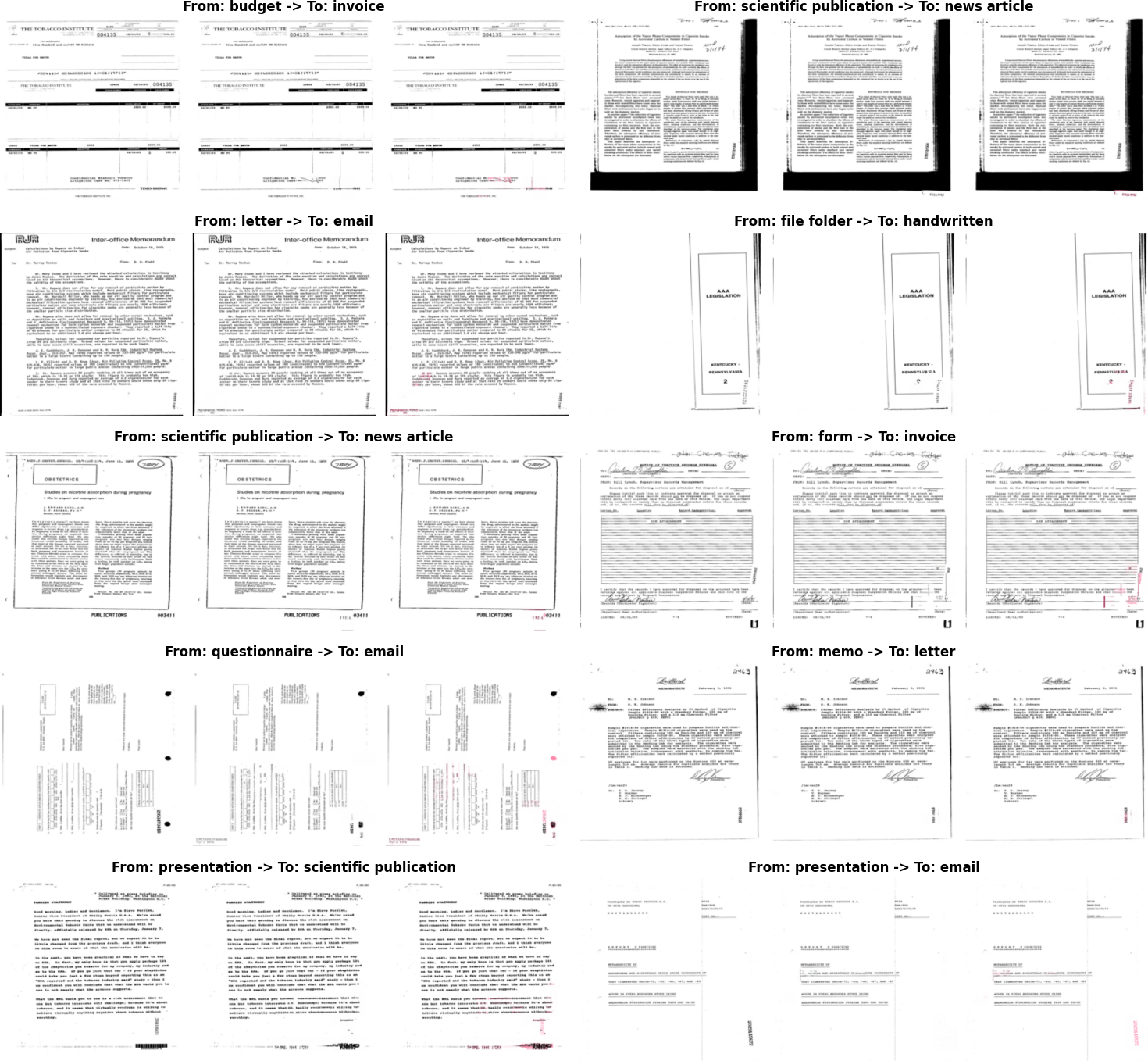}
    \caption{Examples of counterfactuals generated on the \rvlcdip{} dataset using ConvNeXt-B~\cite{resnet}, solely through the insertion or deletion of document identification numbers within the documents. From left to right: original image, counterfactual image, and difference map.}
    \label{fig:spurious_corr_rvlcdip}
\end{figure}

While analyzing the results on the \rvlcdip{} and Tobacco3482 datasets, we observed a strong dependence of counterfactual generation on the insertion or deletion of various types of document identification numbers within the documents. In some cases, class predictions were flipped solely due to the insertion or deletion of these numbers. A few examples showing this behavior on the RVL-CDIP dataset are provided in Fig.~\ref{fig:spurious_corr_rvlcdip}.
Note that these observations align with the findings of a previous work~\cite{Saifullah2023-docxai}, which used feature-attribution methods to identify the same spurious correlation between document identification numbers and different document classes in these datasets.
\section{Additional qualitative results}
\label{app:qualitative-results}
\begin{figure}[H]
    \centering
    \includegraphics[width=\textwidth]{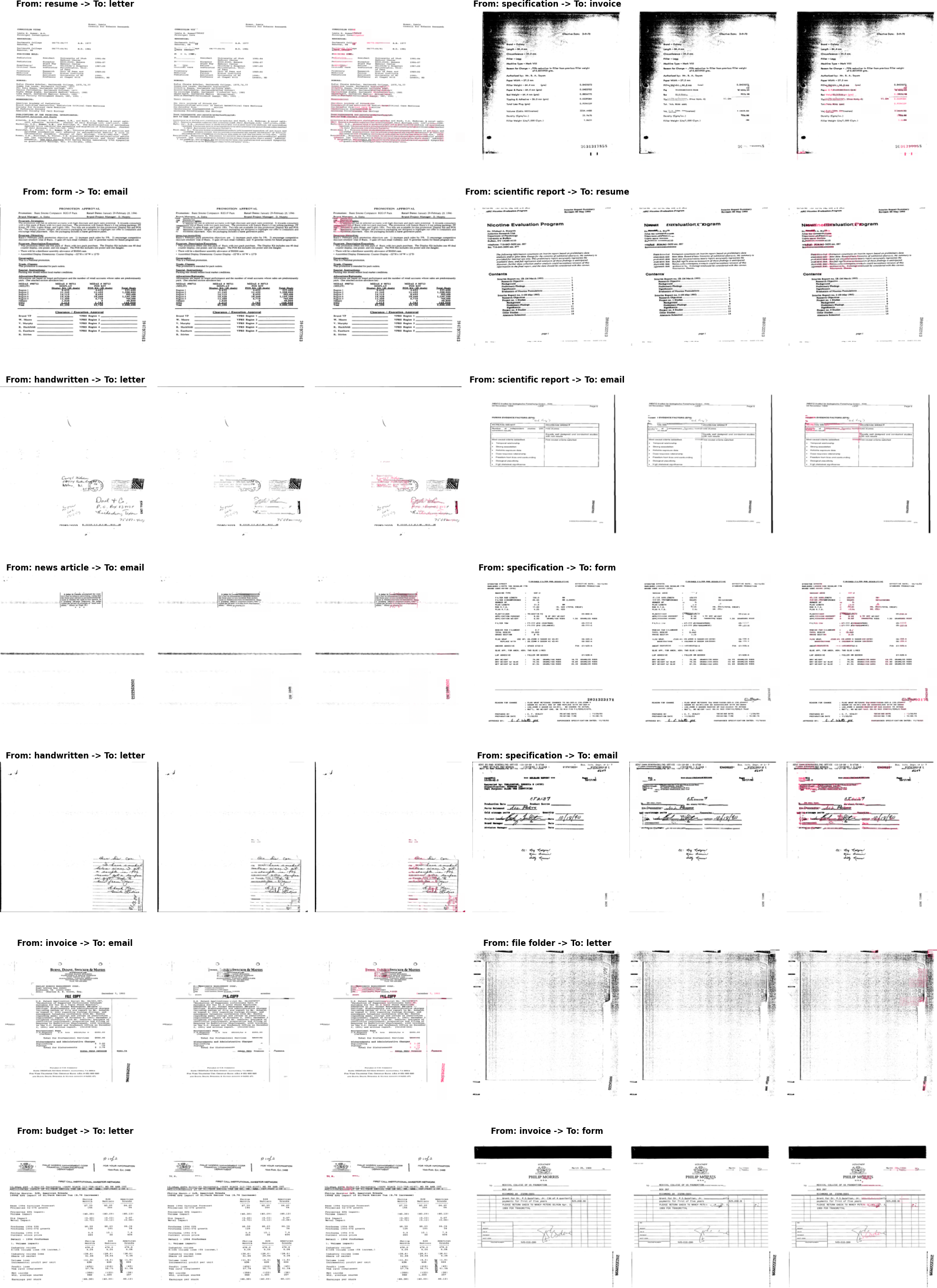}
    \caption{Additional qualitative results for DocVCE on the \rvlcdip{} dataset with DiT-B~\cite{doc-vit} model. From left to right: original image, counterfactual image, difference map.}
    \label{fig:rvlcdip_dit}
\end{figure}

\begin{figure}[H]
    \centering
    \includegraphics[width=\textwidth]{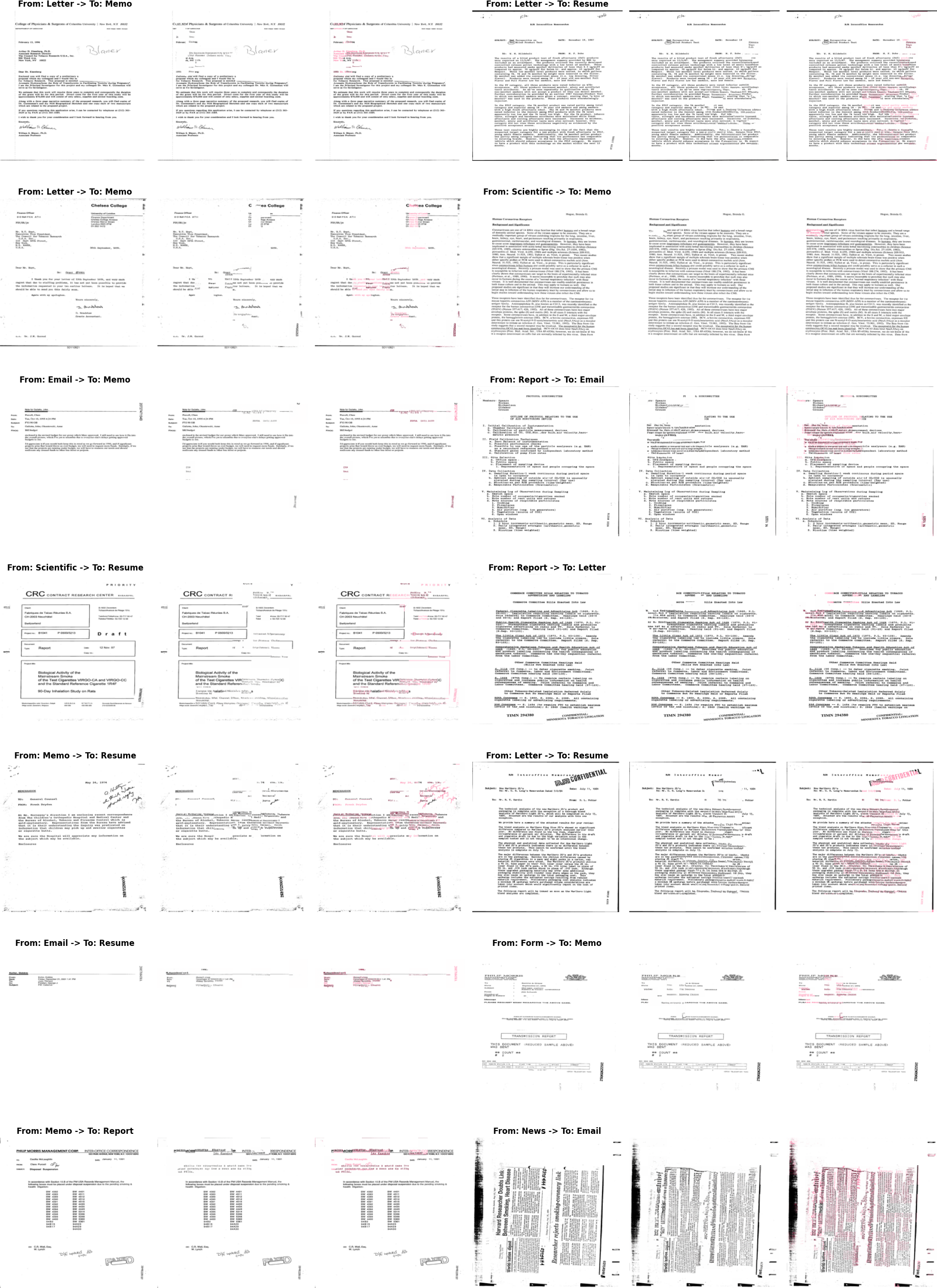}
    \caption{Additional qualitative results for DocVCE on the Tobacco3482 dataset with ConvNeXt-B~\cite{convnext} model. From left to right: original image, counterfactual image, difference map.}
    \label{fig:tobacco3482_convnext_b}
\end{figure}

\begin{figure}[H]
    \centering
    \includegraphics[width=\textwidth]{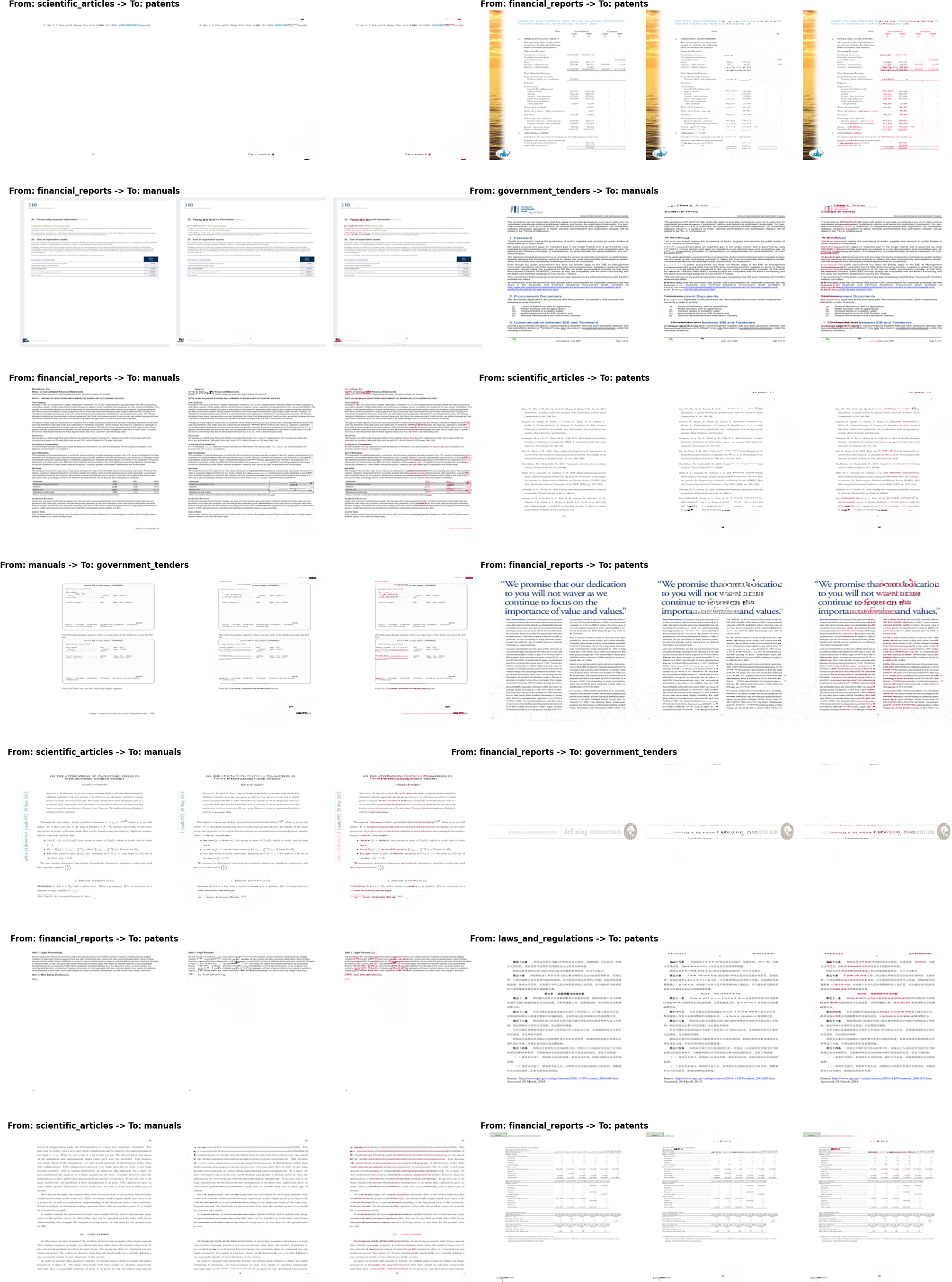}
    \caption{Additional qualitative results for DocVCE on the \doclaynet{} dataset with ResNet-50~\cite{resnet} model. From left to right: original image, counterfactual image, difference map.}
    \label{fig:doclaynet_resnet50}
\end{figure}

In Figures~\ref{fig:rvlcdip_dit}, \ref{fig:tobacco3482_convnext_b}, and \ref{fig:doclaynet_resnet50}, we provide additional qualitative results for the DocVCE approach on the \rvlcdip{} dataset with DiT-B~\cite{doc-vit}, the Tobacco3482 dataset with ConvNeXt-B~\cite{convnext}, and the \doclaynet{} dataset with ResNet-50~\cite{resnet}, respectively.
\begin{figure}[H]
    \centering
    \includegraphics[width=\textwidth]{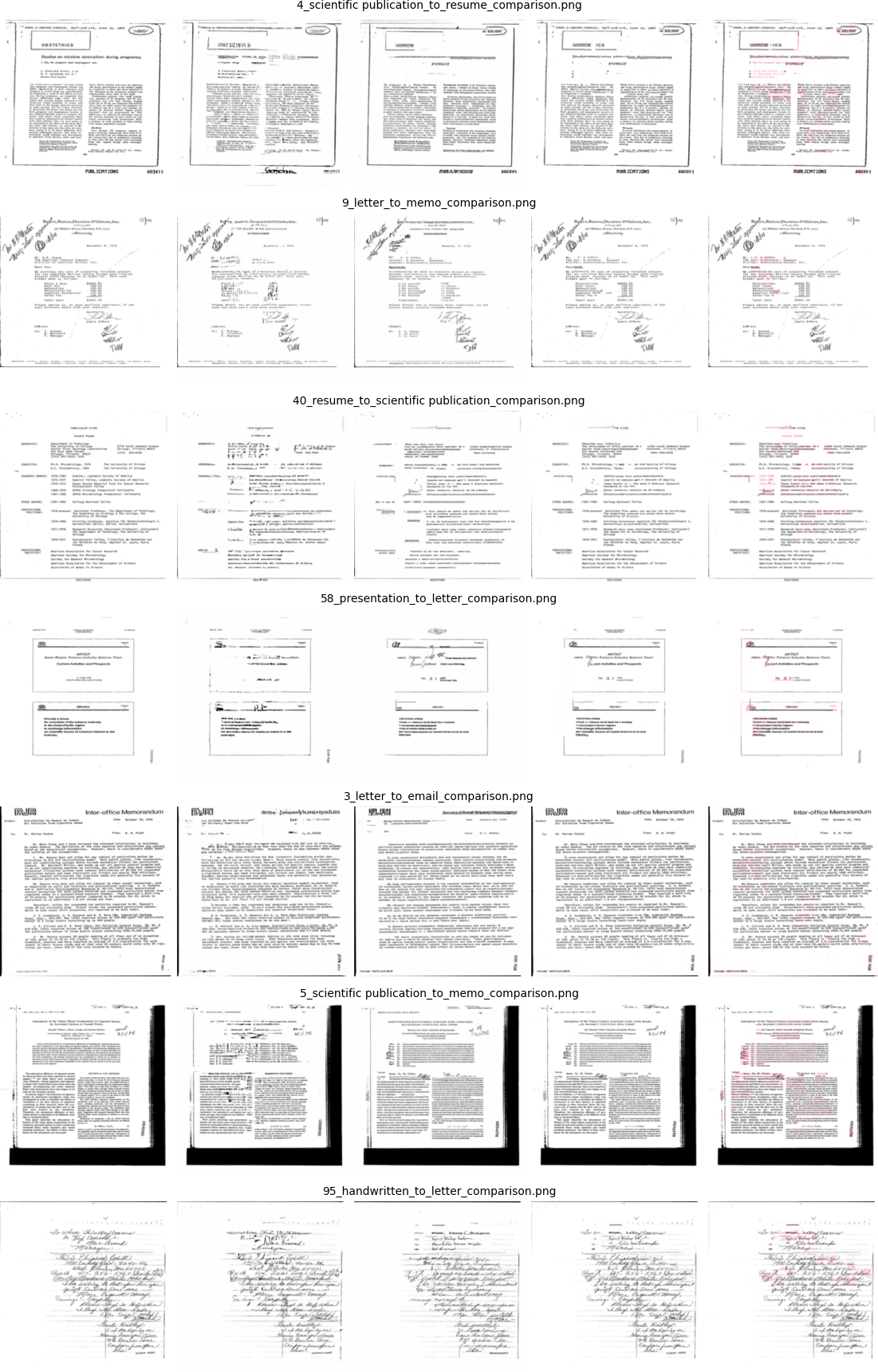}
    \caption{Additional qualitative comparison between the different approaches on \rvlcdip{} dataset with ResNet-50 model. Left to right: original image, LDCE~\cite{farid2023latentdiffusioncounterfactualexplanations}, S-LDCE (ours), DocVCE (ours), DocVCE difference map (ours) From left to right: original image, counterfactual image, difference map.}
    \label{fig:add_qual_rvlcdip_resnet50}
\end{figure}
\begin{figure}[H]
\centering
\includegraphics[width=\textwidth]{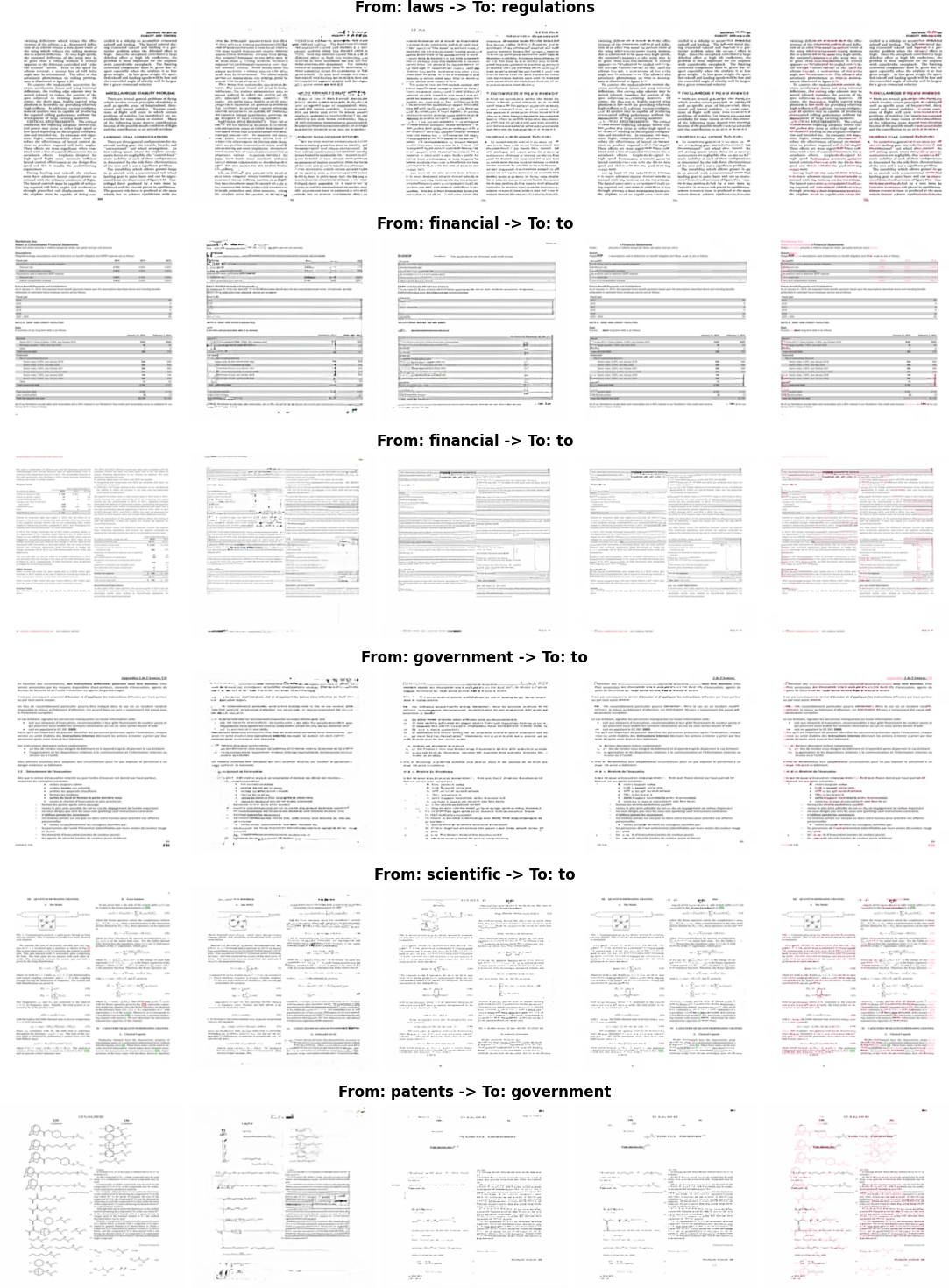}
\caption{Additional qualitative comparison between the different approaches on \doclaynet{} dataset with DiT-B~\cite{doc-vit} model. Left to right: original image, LDCE~\cite{farid2023latentdiffusioncounterfactualexplanations}, S-LDCE (ours), DocVCE (ours), DocVCE difference map (ours) From left to right: original image, counterfactual image, difference map.}
\label{fig:add_qual_doclaynet_dit}
\end{figure}
\begin{figure}[H]
\centering
\includegraphics[width=\textwidth]{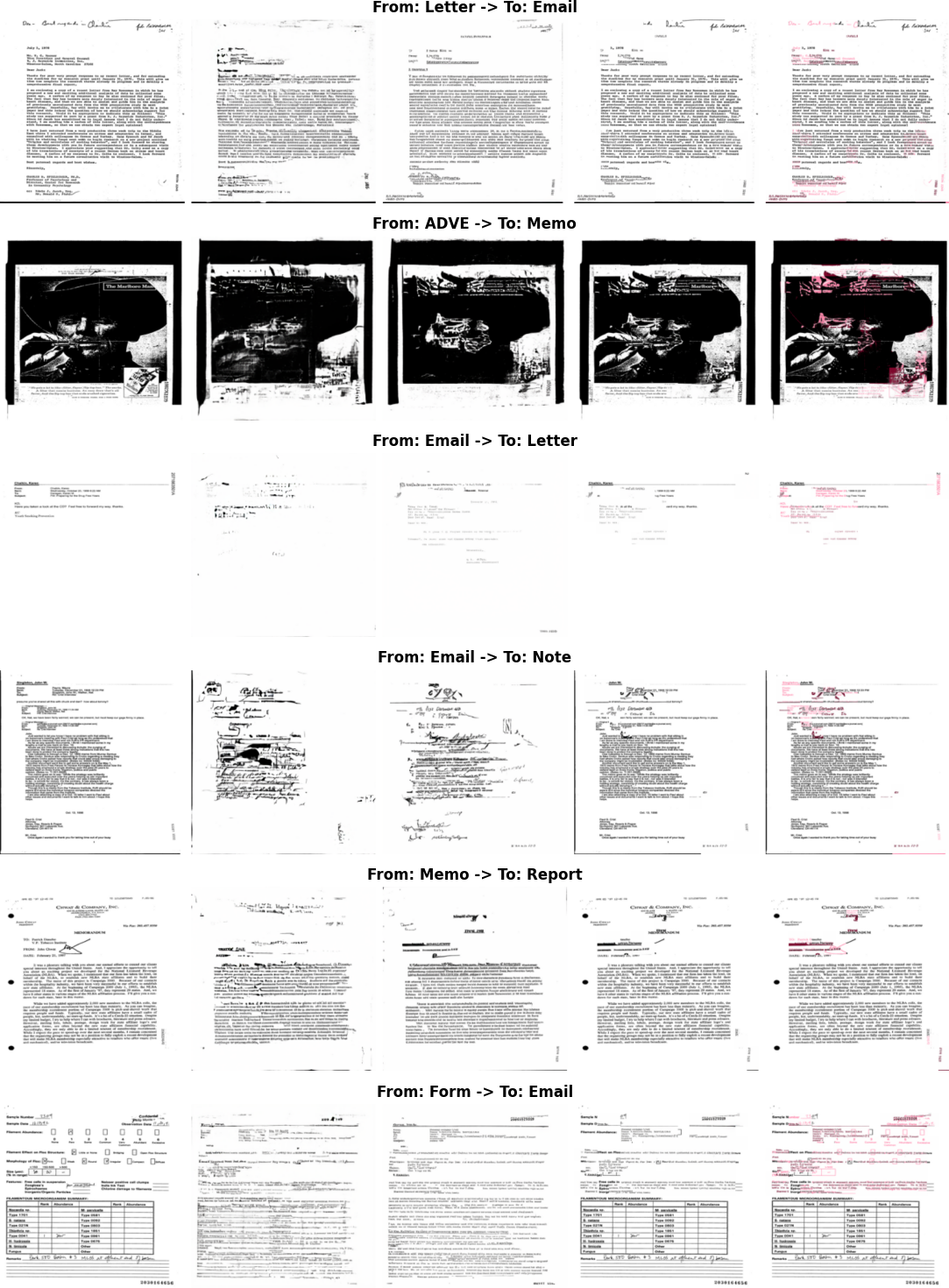}
\caption{Additional qualitative comparison between the different approaches on Tobacco3482 dataset with ConvNeXt-B~\cite{convnext} model. Left to right: original image, LDCE~\cite{farid2023latentdiffusioncounterfactualexplanations}, S-LDCE (ours), DocVCE (ours), DocVCE difference map (ours) From left to right: original image, counterfactual image, difference map.}
\label{fig:add_qual_tobacco_convnext}
\end{figure}
\section{Additional results for qualitative comparison}
In Figures \ref{fig:add_qual_rvlcdip_resnet50}, \ref{fig:add_qual_doclaynet_dit}, and \ref{fig:add_qual_tobacco_convnext}, we provide additional qualitative results for a comparison between different approaches on the \rvlcdip{} dataset with the ResNet-50~\cite{resnet} model, the \doclaynet{} dataset with the DiT-B~\cite{doc-vit}, and the Tobacco3482 dataset with the ConvNeXt-B~\cite{convnext} model, respectively.
\label{app:additional-qualitative-comparison}
\end{document}